\documentclass[a4paper,fleqn]{cas-dc}

\usepackage[authoryear,longnamesfirst]{natbib}

\usepackage{csquotes}
\usepackage{amsmath,amsfonts,amssymb,bm}
\usepackage{nicefrac}
\usepackage{algorithm,algpseudocode}
\usepackage{color}
\usepackage{indentfirst}

\usepackage{graphicx}
\usepackage{fancyhdr,lastpage}
\usepackage{longtable}
\usepackage{makeidx}
\usepackage{multirow}
\usepackage{dcolumn}
\usepackage{epstopdf}
\usepackage{url}
\usepackage{listings}

\usepackage{isotope}
\usepackage{verbatim}
\usepackage{xcolor}
\usepackage{tcolorbox}

\usepackage{float}
\usepackage{stfloats}
\usepackage{subfig}
\usepackage{threeparttable}
\usepackage[export]{adjustbox}

\makeatletter
\renewcommand*\env@matrix[1][\arraystretch]{%
  \edef\arraystretch{#1}%
  \hskip -\arraycolsep
  \let\@ifnextchar\new@ifnextchar
  \array{*\c@MaxMatrixCols c}}
\makeatother

\usepackage{tikz}
\usepackage{pgfplots}
\pgfplotsset{compat=1.14}
\usetikzlibrary{backgrounds,arrows,automata,shapes,positioning,calc,through,spy,shapes,shapes.geometric,shapes.multipart,fit,patterns,fadings,decorations.pathreplacing,calligraphy}
\pgfdeclarelayer{background}
\pgfdeclarelayer{foreground}
\pgfsetlayers{background, main, foreground}

\usepackage{siunitx}
\DeclareSIUnit \parsec {pc}
\DeclareSIUnit \electronvolt {eV}
\DeclareSIUnit \pixel {px}
\DeclareSIUnit \arcmin {arcmin}
\DeclareSIUnit \erg {erg}
\DeclareSIUnit \joul {J}

\usepackage[nolist]{acronym}
\usepackage{pifont}
\newcommand{\cmark}{\ding{51}}%
\newcommand{\ccmark}{\ding{52}}%
\newcommand{\xmark}{\ding{55}}%

\newcommand{\reftab}[1]{Table~\ref{#1}}
\newcommand{\refeq}[1]{\eqref{#1}}

\usepackage{hyperref}

\newlength{\subcolumnwidth}
\newenvironment{subcolumns}[1][0.45\columnwidth]
 {\valign\bgroup\hsize=#1\setlength{\subcolumnwidth}{\hsize}\vfil##\vfil\cr}
 {\crcr\egroup}
\newcommand{\nextsubcolumn}[1][]{%
  \cr\noalign{\hfill}
  \if\relax\detokenize{#1}\relax\else\hsize=#1\setlength{\subcolumnwidth}{\hsize}\fi
}
\newcommand{\nextsubfigure}{\vfill}

\newcommand{\minus}{\scalebox{0.75}[1.0]{$-$}}

\definecolor{amber}{rgb}{1.0, 0.75, 0.0}

\def\tsc#1{\csdef{#1}{\textsc{\lowercase{#1}}\xspace}}
\tsc{WGM}
\tsc{QE}
\newcommand{\PREPRINTYEAR}{2024}
\newcommand{\PUBLISHEDIN}{Ocean Engineering}
\newcommand{\DOI}{https://doi.org/10.1016/j.oceaneng.2024.119164} 

\usepackage[placement=top,vshift=-7]{background}
\SetBgScale{1.0}
\SetBgContents{\PUBLISHEDIN. PREPRINT VERSION - DO NOT DISTRIBUTE. \href{https://doi.org/\DOI}{DOI \DOI}}
\SetBgColor{black}
\SetBgAngle{0}
\SetBgOpacity{1.0}

\begin{document}

\thispagestyle{empty}
\onecolumn
{
  \topskip0pt
  \vspace*{\fill}
  \centering
  \LARGE{%
    \copyright{} \PUBLISHEDIN~\PREPRINTYEAR.\\
    This manuscript version is made available under the CC-BY-NC-ND 4.0 license \url{https://creativecommons.org/licenses/by-nc-nd/4.0/}.}
    \vspace*{\fill}
}
\NoBgThispage
\twocolumn          	
\BgThispage

\let\WriteBookmarks\relax
\def\floatpagepagefraction{1}
\def\textpagefraction{.001}

\shorttitle{Model predictive control-based trajectory generation for agile landing of unmanned aerial vehicle on a moving boat}    

\shortauthors{O. Prochazka et al.}

\title[mode = title]{Model predictive control-based trajectory generation for agile landing of unmanned aerial vehicle on a moving boat}  

\tnotemark[1,2]

\tnotetext[1]{This work has been supported by the Technology Innovation Institute - Sole Proprietorship LLC, UAE, under the Research Project Contract No. TII/ARRC/2055/2021, CTU grant no SGS23/177/OHK3/3T/13 and the Czech Science Foundation (GAČR) under research project No. 23-06162M.
}
\tnotetext[2]{Multimedia materials: \url{https://mrs.fel.cvut.cz/index.php?option=com_content&view=article&id=246}}

\author[1]{Ond\v{r}ej Proch\'{a}zka}[type=editor,
    auid=000,
    bioid=1,
    orcid=0009-0009-2224-750X]
\cormark[1] 
\ead{prochon4@fel.cvut.cz} 
\ead[url]{https://mrs.fel.cvut.cz/members/phdstudents/ondrej-prochazka}

\author[1]{Filip Nov\'{a}k}[type=editor,
    auid=000,
    bioid=1,
    orcid=0000-0003-3826-5904]

\author[1]{Tom\'{a}\v{s} B\'{a}\v{c}a}[type=editor,
    auid=000,
    bioid=1,
    orcid=0000-0001-9649-8277]

\author[1]{Parakh M. Gupta}[type=editor,
    auid=000,
    bioid=1,
    orcid=0000-0002-6481-2281]

\author[1]{Robert P\v{e}ni\v{c}ka}[type=editor,
    auid=000,
    bioid=1,
    orcid=0000-0001-8549-4932]

\author[1]{Martin Saska}[type=editor,
    auid=000,
    bioid=1,
    orcid=0000-0001-7106-3816]




\address[1]{Czech Technical University in Prague, Faculty of Electrical Engineering, Czechia}

\cortext[1]{Corresponding author} 

\begin{abstract}
This paper proposes a novel trajectory generation method based on \ac{MPC} for agile landing of an \ac{UAV} onto an \ac{USV}’s deck in harsh conditions. 
The trajectory generation exploits the state predictions of the \ac{USV} to create periodically updated trajectories for a multirotor \ac{UAV} to precisely land on the deck of a moving \ac{USV} even in cases where the deck’s inclination is continuously changing. 
We use an \ac{MPC}-based scheme to create trajectories that consider both the \ac{UAV} dynamics and the predicted states of the \ac{USV} up to the first derivative of position and orientation. 
Compared to existing approaches, our method dynamically modifies the penalization matrices to precisely follow the corresponding states with respect to the flight phase. 
Especially during the landing maneuver, the \ac{UAV} synchronizes attitude with the \ac{USV}’s, allowing for fast landing on a tilted deck. 
Simulations show the method's reliability in various sea conditions up to \textit{Rough sea} (wave height \SI{4}{\meter}), outperforming state-of-the-art methods in landing speed and accuracy, with twice the precision on average. 
Finally, real-world experiments validate the simulation results, demonstrating robust landings on a moving \ac{USV}, while all computations are performed in real-time onboard the \ac{UAV}.
\end{abstract}



\begin{keywords}
Model Predictive Control \sep
Unmanned Surface Vehicle \sep
Unmanned Aerial Vehicle \sep
Multi-robot Systems \sep
Landing \sep
Harsh environment
\end{keywords}

\maketitle


\section{INTRODUCTION}
    There has been a significant increase in the utilization of \acfp{UAV} in various domains in recent years, including tasks above water surfaces.
    \acp{UAV} have proven beneficial in marine wildlife monitoring \cite{fortuna2013using, yang2022uav}, search and rescue operations~\cite{SearchandRescue2}, and offshore safety monitoring to identify swimmers in distress \cite{Sharma2022}. 
    Moreover, \acp{UAV} can be deployed to detect garbage on the water surface \cite{su141811729}.
    Individual garbage pieces can be then picked up by the drone itself, or its position can be shared from the \ac{UAV} to the \acf{USV} for collection \cite{8929537}.
    As it is clear, not only, from the aforementioned literature, the \acp{UAV} are a valuable resource for operations and study of maritime environments.
    However, their limited operation time necessitates a carrier ship to transport the \ac{UAV} to the mission location for takeoff and landing in the vicinity of a mission area.
    \begin{figure}[!t]
      \centering
      \input{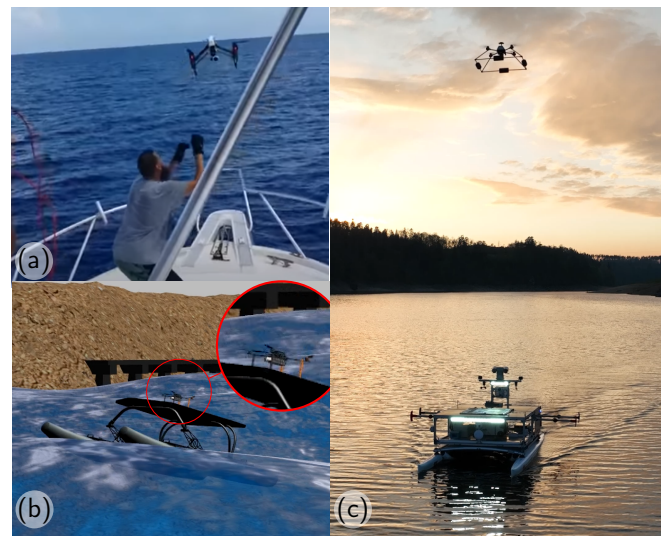}
      \caption{Motivation (a) manual landing of the UAV in the real world \cite{DailyPicksandFlicks}, (b) autonomous landing of the UAV in simulation in the high wave environment, and (c) photo of the UAV together with the USV research platform deployed in the real world.}
      \label{fig:intro_figure}
      \vspace{-0.7cm}
    \end{figure}  
    The carrier ship brings the additional advantage of having better payload capacity and energy storage, and thus, can also be used to recharge the \ac{UAV} \cite{en10060803}.
    To maintain mission efficiency and eliminate the need for human intervention, the landing needs to be autonomous, which is challenging due to the severe conditions in a maritime environment.
    It is estimated that the probability of witnessing waves with a height of half a meter or more is approximately \SI{90}{\percent} worldwide~\cite{fossen2011handbook}.
    Waves affect ship's position, altitude, attitude, and velocity such that the \ac{USV} is sliding, tilting, and heaving simultaneously, making \ac{UAV} landings challenging, especially on small \acp{USV}, preventing the use of standard landing maneuvers.
    Available mission time is further diminished by the time required for a visual search of the \ac{USV} for landing, and additionally, for a collaborative mission (e.g. collaborative object manipulation on the water surface \cite{novak2024collaborative}), it is reasonable to assume that communication exists between the two vehicles. 
    Therefore, 
    this communication link is used to also share the position of the \ac{USV}, allowing for a direct and time-efficient approach to the location of the \ac{USV}. 
    At the same time, we aim to shorten the landing maneuver itself as much as possible. 
    While possibly dangerous in harsh sea conditions, the direct return to \ac{USV} and the agile landing can significantly prolong the time for the main mission objectives.

    This paper introduces a novel approach that enables \acp{UAV} to land autonomously on a moving or rocking \ac{USV} in harsh environments utilizing a \acf{MPC}-based trajectory generation in the process.
    This removes human intervention (see \autoref{fig:intro_figure}) and increases the safety and reliability of the \ac{UAV}'s usage on the high seas.
    The proposed method relies on accurate estimation and prediction of \ac{USV}'s states, including position, rotation, and their first derivative, which are obtained from multiple sources to allow for landing in low visibility or harsh sea conditions \cite{tro2023novak}.
    In contrast to our approach, most of the state-of-the-art methods \cite{xu2020vision, abujoub2019development, 6404893, drones2020015, keller2022robust, persson2019model} rely on simplified models of the \ac{USV}, which does not guarantee precise estimation of the \ac{USV}'s states.
    Consequently, landing accuracy in harsh environments cannot be assured due to the missing information about the \ac{USV}'s movement, which could result in an unsafe procedure or even a crash.
    To this end, our trajectory generator incorporates significantly more information about the \ac{USV} state that is available during the individual phases of the \ac{UAV}'s operation and adjusts penalization matrices' values that prioritize specific states for tracking.
    The information about roll, pitch, yaw, linear movements, and Euler rates is taken into account to increase precision and reduce the harshness of the touchdown.
    This enables a significant increase in the performance of landings in various sea conditions and even on an inclined \ac{USV} deck.

    The simulation results demonstrate the proposed system's reliable performance in landing the \ac{UAV} on a small scale \ac{USV} under \textit{Moderate sea} and even \textit{Rough sea} conditions in which the height of the waves reaches up to \SI{4}{\meter}. 
    The system underwent extensive testing in more than 1,000 simulation experiments and was compared to the latest state-of-the-art work \cite{gupta2022landing}.
    We show approximately two times better touchdown accuracy with a \SI{50}{\percent} faster landing maneuver on average.
    It is worth mentioning that our method achieved a success rate of \SI{100}{\percent} in most tested environment setups of the different sea conditions, with only two particularly challenging setups showing a slightly lower success rate, resulting in aborted and repeated landing.
    Moreover, the system was repeatedly tested in a real-world environment where its robustness and performance indicated in simulations were confirmed.

\section{RELATED WORKS}
    Landing on a rocking and heaving platform, especially on a marine vessel in bad weather conditions, is a very challenging maneuver in which the helicopter can crash into the vessel or fall from the deck into the water.
    Even highly trained military pilots have difficulty with such a maneuver and thus rely on the visualization of horizon reference \cite{stingl1970vtol} or use beartrap for a safer landing.
    Since it is impossible to use beartrap for small \acp{UAV}, the authors in \cite{XU2024117306} proposed a manipulator-assisted landing system that captures the hovering \acp{UAV} using the tethered connection.
    The landing is done on a \ac{USV}'s motion compensation platform, which is unlike the platform mentioned in the work of \cite{TIAN2024118223} capable of compensating pitch and roll motions of the \ac{USV}.
    The presented systems require a multiple-joint manipulator placed on the \ac{USV}, and even so, for the \ac{UAV}'s position control, visual servoing feedback is used.
    
    The visual servoing approaches can be divided into two categories: \ac{PBVS} and \ac{IBVS} \cite{9739493}.
    The \ac{PBVS} method works in Cartesian space, whereas the \ac{IBVS} approach works directly with the image input from sensors.
    The \ac{IBVS} suffers from camera calibration errors, although the \ac{PBVS} may suffer from position estimation errors.
    Moreover, it is noteworthy that the \ac{IBVS} method \cite{9739493} assumes that the \ac{UAV}'s attitude changes are small, and can thus be neglected.
    Furthermore, the \ac{IBVS} method uses a prediction of the landing zone for smoother control by using a Kalman filter \cite{kalman1960new}.
    The solution described in \cite{9739493} was tested on a simulated \ac{USV} deck in the real world, proving that the \ac{UAV} can detect and land on a tilting deck.
    Also, the dynamic landing algorithm presented in \cite{ding2023research}, which enables landing on the heaving \ac{USV}, was tested in real-world scenarios. 
    However, the authors do not present the roll and pitch tracking results upon the touchdown when \ac{USV} is rocking.
    
    The \ac{PBVS} based on \ac{PID} controller introduced in \cite{xu2020vision} was used for vision-based autonomous landing on a boat where an AprilTag \cite{olson2011tags} was placed on the helipad.
    The visual slide landing is suitable in a situation where it is not possible to hover above the helipad and perform a regular landing because of sails or cables \cite{keller2022robust}. 
    Another work demonstrating the capability of landing on a one-direction fast-moving platform is presented in \cite{zhang2024precise}.
    These landing approaches mostly do not model the \ac{USV}'s roll, pitch, and heave motions. 
    
    The aforementioned visual-based approaches do not work well in limited lighting conditions.  
    To enable landing at night, the marine vessel's deck can be equipped with infrared markers \cite{meng_visualinertial_2019} that are used to navigate fixed-wing aircraft to the runway during autonomous landing. 
    Other approaches equip the \ac{UAV} with \ac{lidar} sensors \cite{ ABUJOUB2020106169, 8604820}.
    \ac{lidar} can detect the vessel's movements and use it for the \ac{SPA} \cite{8604820}.
    This algorithm includes mode detection, estimation, and prediction using \acl{FFT} to identify the frequency, amplitude, and phase of the roll and pitch movements. 
    Another approach that can determine a proper time for landing is based on estimating the ship's energy using \ac{LPI}, where the pertinent motions are roll, pitch, and heave~\cite{ABUJOUB2020106169}.
    Regarding the vertical oscillations, the \ac{AHC} algorithm is proposed in the same paper \cite{ABUJOUB2020106169}.
    The approach of the \ac{AHC} algorithm is to stabilize the \ac{UAV} over the deck to ensure a landing that considers the raising and lowering of the deck. 
    Therefore, it can be used together with \ac{SPA} or \ac{LPI}.

    The drawback of the methods mentioned above is that they are purely vision-based. 
    As such, if the vessel is not in the \ac{FOV} of the \ac{UAV}'s sensors, the global position of the \ac{USV} is unknown.
    The \ac{UAV} must find the \ac{USV} by searching the area, which is an energy-consuming procedure.
    Furthermore, it is anticipated that cooperation between the \ac{UAV} and \ac{USV} will be required for certain mission tasks, and therefore the \ac{USV} is expected to be in the vicinity of the \ac{UAV}.
    A wireless communication between the \ac{UAV} and the \ac{USV} for sharing the global position of the \ac{USV} is presented in \cite{6404893, paris2020dynamic}.
    Nevertheless, it is shown that the \ac{GNSS} precision is not sufficient for small \acp{UAV} and mainly for the altitude estimation.  
    Due to the potential unreliability and imprecision of \ac{GNSS}, we rely on a method that combines \ac{UAV} onboard vision-based sensors with \ac{USV} onboard sensors to accurately estimate the states of the \ac{USV} \cite{tro2023novak}.
    The method proposed in \cite{tro2023novak} utilizes data from the \ac{USV}'s \ac{GNSS} for a rough estimation of its location and incorporates visual relative localization system \ac{UVDAR} \cite{uvdd1} and visual fiducial system AprilTag \cite{Wang2016}.
    Thanks to the \ac{UVDAR} system, localization of the \ac{USV} can be done even in reduced visibility conditions, such as evening or night.
    To enhance the precision of the \ac{USV}'s orientation and angular rate estimation and prediction during the landing maneuver, data from the \ac{USV}'s \ac{IMU} is used.

    \begin{table}[t!]
        \centering
        \caption{Comparison of the aspects of landing approaches for the vessels in related research works.\\\hspace{\textwidth}}
        \label{tab:comparison_stoa}
        \vspace{-0.4cm}
    \begin{threeparttable}
        \begin{tabular}{lcccccc}
        \hline
        \multicolumn{1}{l}{Work} &
         \begin{tabular}[c]{@{}c@{}} Dp\tnote{1} \end{tabular} &
         \begin{tabular}[c]{@{}c@{}} Al\tnote{2} \end{tabular}  &
          \begin{tabular}[c]{@{}c@{}}Hc\tnote{3}\end{tabular} &
          \begin{tabular}[c]{@{}c@{}}6-DOFs\\ USV model \end{tabular} &
          \begin{tabular}[c]{@{}c@{}}Real-world\\  deployment\end{tabular} \\ \hline \hline
          (1) 
        & \xmark & \xmark & \xmark  & \xmark & \cmark \\
        (2) 
        & \xmark & \xmark & \cmark  & \cmark & \xmark  \\ 
        (3)  
        & \xmark  & \xmark & \xmark  & \xmark & \xmark \\ 
        (4) 
        & \xmark & \cmark & \cmark  & \xmark & \cmark  \\ 
        (5) 
        & \xmark & \xmark & \cmark  & \cmark & \xmark  \\ 
        (6) 
        & \xmark & \xmark & \xmark  & \xmark & \cmark \\ 
        (7) 
        & \xmark  & \xmark & \xmark  & \xmark & \xmark  \\ 
        (8) 
        & \xmark  & \xmark &  \xmark  & \xmark & \cmark \\ 
        (9) 
        & \xmark & \xmark & \cmark  & \cmark & \cmark \\ 
        (10) 
        & \xmark & \cmark & \cmark  & \cmark & \cmark \\ 
        (11)  
        & \xmark & \cmark & \cmark  & \cmark & \xmark \\ 
        \textbf{This work}                                                          & \ccmark  & \ccmark & \ccmark  & \ccmark & \ccmark \\ \hline
        \end{tabular}
        \begin{tablenotes}
            \item[-] (1) \cite{6404893}, (2) \cite{doi:10.2514/6.2014-1298}, (3) \cite{drones2020015}, (4) \cite{persson2019model}, (5) \cite{ABUJOUB2020106169}, (6) \cite{xu2020vision}, (7) \cite{9739493}, (8) \cite{keller2022robust}, (9) \cite{gupta2022landing}, (10) \cite{ding2023research}, (11) \cite{stephenson2024distributed}
            \item[1] Dynamic adjustments of the penalization matrices to achieve better tracking of the individual USV's states during the flight and especially during the final touchdown. 
            \item[2] UAV aligns its position and orientation up to the first derivative with USV's upon touchdown.
            \item[3] Heave motion compensation.
        \end{tablenotes}
    \end{threeparttable}
    \vspace{-0.4cm}
    \end{table}

    To increase the precision of the \ac{UAV}'s control algorithm, the \ac{PID} controller is nowadays being replaced by a method referred to as \ac{MPC}. 
    This is also known as receding horizon control or moving horizon control \cite{garcia1989model, 845037}.
    In \cite{doi:10.2514/6.2014-1298}, nonlinear \ac{MPC} is used to control the landing of a helicopter on a navy vessel based on a prediction of the quiescent periods, which is made based on long-term observation of the \ac{USV}'s movements.
    This system was outperformed by methods presented in \cite{gupta2022landing, stephenson2024distributed}.
    Both methods use the same approach for a landing procedure, waiting for a low tilt of the \ac{USV}. 
    The landing maneuver is performed based on the defined internal landing cost function incorporated into the \ac{MPC}, which penalizes the \ac{USV}'s rocking movements.
    Both \cite{doi:10.2514/6.2014-1298, gupta2022landing} landing approaches may require a long time for safe landing.
    Considering the limited flight time of \acp{UAV}, it is crucial to optimize the landing maneuver and shorten the overall return-to-home procedure. 
    Thus, the maximum flight time can be used for the \ac{UAV}'s primary task (e.g., off-shore infrastructure inspection, oil spill monitoring, marine life tracking, and vessel inspection).
    However, these approaches \cite{doi:10.2514/6.2014-1298, gupta2022landing} need to wait for the \ac{USV}'s deck to be fully leveled or for calmness in wave motion, assuming that the surge and sway motions of the \ac{USV} are negligible.
   
    In the proposed approach, the \ac{UAV} compensates the \ac{USV}'s attitude by synchronizing the \ac{UAV}'s attitude with the \ac{USV}'s, allowing for fast landing on a tilted deck. 
    Furthermore, \ac{USV} model, which is used, includes wave modeling that we introduced in \cite{tro2023novak}, contrasting with the state-of-the-art approaches, where such information is not considered.
    The model of the waves is crucial for heave motion estimation \cite{reis2023discrete} and prediction.
    Such possible heave compensation integrated in the landing technique ensures low impact velocity during touchdown.
    These features make it usable in varying sea conditions and more efficient for safe and fast mission completion, even on \textit{Rough sea}.
    A comparison summary of our proposed system with existing state-of-the-art approaches is shown in \reftab{tab:comparison_stoa}.
    According to this table, many works utilize the full 6 \acp{DOF} \ac{USV} model, heave motion compensation, and take into account \ac{USV}'s attitude angles.
    While many approaches have been tested in real-world scenarios, none incorporate dynamic penalization matrices adjustments to better track the \ac{USV}'s individual states during the final touchdown.

\section{VEHICLES' MODELS}\label{sec:usv_uav_models}
\begin{figure}[b!]
      \centering
      \includegraphics[width=0.43\textwidth]{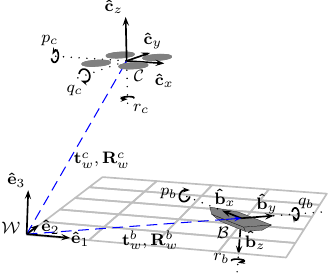}
      \caption{Illustration of the UAV's and USV's coordinate systems, where $\mathcal{W}$ represents local ENU frame, $\mathcal{C}$ denotes the multirotor's body-fixed frame, and $\mathcal{B}$ symbolizes the USV's body-fixed frame. 
      The rotation matrix from the world frame to the USV’s body-fixed frame is represented as $\mathbf{R}_{w}^{b}$, and the translation vector is denoted as $\mathbf{t}_{w}^{b}$.
      The rotation matrix $\mathbf{R}_{w}^{c}$ and translation vector $\mathbf{t}_{w}^{c}$ denotes rotation and translation from the world frame to the UAV’s body-fixed frame.
      The angular velocity of the UAV and USV is denoted with $(p, q, r)$ with the appropriate subscripts.}
      \label{fig:coordinate_system}
    \end{figure}

    This section introduces the mathematical model and physics background of our work.
    The mathematical model of the \ac{USV} serves as the basis for estimating and predicting \ac{USV}'s states.
    This section further comprises the derivation of the \ac{UAV}'s mathematical model, which is crucial for the subsequent MPC-based trajectory generation. 
    To be able to describe the models, let us define the local \ac{ENU} coordinate frame $\mathcal{W}$ together with the \ac{UAV}'s body coordinate frame ($\mathcal{C}$) and \ac{USV}'s coordinate frame ($\mathcal{B}$) as shown in \autoref{fig:coordinate_system}.

    The precise modeling of the \ac{USV} is crucial for accurate estimation and prediction of its states, which are essential conditions for designing a safe and accurate landing maneuver.
    For this purpose, the 6 \ac{DOF} model of the \ac{USV} utilized in this work is composed of the standard Fossen's \ac{USV} model \cite{fossen2002marine} and model of the waves as is proposed in \cite{tro2023novak}.
    According to \cite{tro2023novak}, the general definition of the \ac{USV} model looks as follows 
    \begin{align}\label{eq:usv_model}
        \bm{\dot{\eta}}_L &= \bm{\nu}, \\
        \bm{\dot{\nu}} &= -\mathbf{M}^{-1}(\mathbf{D}\bm{\nu} + \mathbf{G}\bm{\eta}_L) + \mathbf{C}_{\text{wave}, \bm{\nu}}\mathbf{{x}}_{\text{wave}, \nu}, \\ \label{eq:usv_model3}
        \mathbf{\dot{x}}_{\text{wave}, \nu} &= \mathbf{A}_{\text{wave}, \bm{\nu}}\mathbf{{x}}_{\text{wave}, \nu},
    \end{align}
    where $\bm{\eta}_L = [\hat{b}_x, \hat{b}_y, \hat{b}_z, \phi_b, \theta_b, \psi_b]^\top \in \mathbb{R}^{3} \times \mathcal{S}^3$ is a vector composed of positions and orientations in \emph{Vessel parallel coordinate system} \cite{fossen2011handbook}.
    Euler angles are $\phi_b, \theta_b$, and $\psi_b$, $\bm{\nu} = [u_b, v_b, w_b, p_b, q_b, r_b]^\top \in \mathbb{R}^{6}$ is a vector that contains the linear and angular speeds for all axes.
    $\mathbf{G}$~represents matrix of gravitational and buoyancy forces, $\mathbf{D}$~describes the system's damping, $\mathbf{M}~=~\mathbf{M}_{RB}+\mathbf{M}_A$ is a sum of the \ac{USV}'s body inertia matrix and hydrodynamics inertia matrix.
    State-space representation matrix of the waves' damping and frequency is denoted with $\mathbf{A}_{\text{wave}, \bm{\nu}}$.
    Matrix $\mathbf{C}_{wave,\nu}$ is the diagonal output matrix of the size, which corresponds to the number of modeled waves, and $\mathbf{{x}}_{\text{wave}, \nu}$ is the state vector, including all states of the individual waves components.
    Notice the model \refeq{eq:usv_model}--\refeq{eq:usv_model3} is used solely for \ac{USV} state estimation and prediction onboard the \ac{UAV}, maintaining precision even in challenging conditions without relying on inputs like rudder and propellers, which would reduce robustness by increasing reliance on communication stability.

    Based on the already formulated \ac{USV}'s model, the full nonlinear model of the \ac{UAV} is defined as a rigid body model in three-dimensional space.
    Generalized positions \refeq{eq:position_vector} and velocities \refeq{eq:velocity_vector} are designated as 
    \small{\begin{align}\label{eq:position_vector}
        \bm{\eta}_c &= [\bm{\eta}_{p_c}^\intercal, \bm{\eta}_{o_c}^\intercal]^\top = [\hat{c}_x, \hat{c}_y, \hat{c}_z, \phi_c, \theta_c,\psi_c]^\top \in \mathbb{R}^{3} \times \mathcal{S}^3, \\
        \bm{\upsilon}_c &= [\bm{\upsilon}_l^\intercal, \bm{\upsilon}_a^\intercal]^\top = [v_x,v_y, v_z, v_\phi, v_\theta, v_\psi]^\top \in \mathbb{R}^{6}, \label{eq:velocity_vector}
    \end{align}}
    where vector $\bm{\eta}_c$ is composed of position $\bm{\eta}_{p_c} = [\hat{c}_x, \hat{c}_y, \hat{c}_z]^\top$ and orientation $\bm{\eta}_o = [\phi_c, \theta_c, \psi_c]^\top$.
    The Euler angles are described in intrinsic \textit{zyx}-convention.
    The vector $\bm{\upsilon}_c$ includes linear velocity $\bm{\upsilon}_l~=~[v_x,v_y, v_z]^\top$ and the Euler rate vector $\bm{\upsilon}_a~=~[v_\phi, v_\theta, v_\psi]^\top$.
    The position $\bm{\eta}_c$ and velocity $\bm{\upsilon}_c$ are defined in the local \ac{ENU} coordinate frame $\mathcal{W}$.

    The translational dynamics of the \ac{UAV} is defined as
    \begin{align}\label{eq:uav_kinematics}
        \bm{\dot{\eta}}_{p_c} &= \bm{\upsilon}_l,\\
        \bm{\Ddot{\eta}}_{p_c}
        &=
        -g\,\mathbf{\hat{e}}_3
        +
        \mathbf{R}_c^w \dfrac{\mathbf{T_C}}{m},
        \label{eq:kinetics_quadcopter_lin_acc}
    \end{align}
    where $g$ denotes gravitation acceleration, $\mathbf{\hat{e}}_3$ is an element of the \emph{standard basis}, and $m$ is the quadcopter weight.
    The $\mathbf{R}_c^w$ is an intrinsic \textit{xyz}-convention Euler rotation matrix.
    The total force (thrust) $\mathbf{T_C}$ is generated in the body-fixed frame as
    \begin{align}\label{eq:copter_total_force}
        \mathbf{T_C} &= \begin{bmatrix} 0 & 0 &
        k_T \sum_{i=1}^{4} \omega_i^2 
        \end{bmatrix}^\top,
    \end{align}
    where $k_T$ is a lift constant for a motor with a propeller and $\omega_i$ is the rotor's angular velocity.
    The angular dynamics in local coordinate frame $\mathcal{W}$ are defined as
    \begin{align}
    \label{eq:angular_velocity}
        \bm{\dot{\eta}}_{o_c} &= \bm{\upsilon}_a, \\
        \bm{\Ddot{\eta}}_{o_c}
        &=
        (\mathbf{T}_c^w \mathbf{I}_c \mathbf{T}_w^c)^{-1}\left({\bm{\tau_C}}  - \mathbf{C_t} \bm{\dot{\eta}}_{o_c}\right),
        \label{eq:kinetics_quadcopter_angular_acc}
    \end{align}
    where $\mathbf{C_t}$ is the Coriolis term, which contains centripetal and gyroscopic terms, as shown in \cite{raffo2010integral}.
    The $\bm{\tau_C}$ stands for the total torque generated by the rotors and $\mathbf{I}_c \in \mathbb{R}^{3x3}$ is an inertial matrix composed of $I_{xx}, I_{yy}, I_{zz}$ on its diagonal. 
    As shown in \cite{Jaln2014NonlinearAC}, the transformation matrix $\mathbf{T}_{w}^c$ = $(\mathbf{T}_{c}^w)^\top$ is defined as
    \begin{align}\label{eq:uav_transform_angular_velocity}
        \mathbf{T}_{w}^{c} = 
        \mathbf{R}^\intercal_{x, \phi_c}\mathbf{R}^\intercal_{y, \theta_c}\mathbf{R}^\intercal_{z, \psi_c}\mathbf{\hat{e}_3} + \mathbf{R}^\intercal_{x, \phi_c}\mathbf{R}^\intercal_{y, \theta_c}\mathbf{\hat{e}_2} + \mathbf{R}^\intercal_{x, \phi_c}\mathbf{\hat{e}_1},
    \end{align}
    where $\mathbf{\hat{e}}_1, \mathbf{\hat{e}}_2, \mathbf{\hat{e}}_3$ are elements of the \emph{standard basis}.
    It is also worth mentioning that the inversion $\mathbf{T}_c^w$ of $\mathbf{T}_{w}^c$ from \refeq{eq:uav_transform_angular_velocity} is not defined for $\theta_c = \frac{\pi}{2} + j\pi, j \in \mathbb{Z}$. 

    In order to provide the linear approximation of the equations of motion that were shown in \refeq{eq:uav_kinematics}, \refeq{eq:kinetics_quadcopter_lin_acc}, \refeq{eq:angular_velocity}, and \refeq{eq:kinetics_quadcopter_angular_acc}, an equilibrium point is needed \cite{da2015benchmark}. 
    We consider as equilibrium the hovering state of the \ac{UAV}, which is defined as a state with zero derivatives of position.
    Thus, the linearized solution is introduced as 
    \small{\begin{align}\label{eq:uav_linear_model}
        \begin{bmatrix}
            \bm{\dot{\eta}}_{p_c} \\
            \bm{\dot{\eta}}_{o_c} \\
            \bm{\ddot{\eta}}_{p_c} \\
            \bm{\ddot{\eta}}_{o_c}
        \end{bmatrix} 
        =
        \underbrace{\begin{bmatrix}
            \mathbf{0}^{6\times 3} & \mathbf{0}^{6\times 3} & \mathbf{I}^{6\times 6} \\
            \mathbf{0}^{6\times 3} & \mathbf{A_p} & \mathbf{0}^{6\times 6}
        \end{bmatrix}}_{\mathbf{Ac}}
        \begin{bmatrix}
            \bm{\eta}_{p_c} \\
            \bm{\eta}_{o_c} \\
            \bm{\dot{\eta}}_{p_c} \\
            \bm{\dot{\eta}}_{o_c} 
        \end{bmatrix} 
        +
        \underbrace{\begin{bmatrix}
            \mathbf{0}^{8\times 4} \\
            \mathbf{B_p}
        \end{bmatrix}}_{\mathbf{Bc}}
        \begin{bmatrix}
            \omega_{1}^2 \\
            \omega_{2}^2 \\
            \omega_{3}^2 \\
            \omega_{4}^2
        \end{bmatrix},
    \end{align}}
    where $\mathbf{A_p}$ is a sparse matrix which contains nonzero elements located at specific positions as follows
    \small{
    \begin{align}
        \mathbf{A_p}^{[1,2]} &= \frac{k_T(\omega^2_{h_{1}} + \omega^2_{h_{2}} + \omega^2_{h_{3}} + \omega^2_{h_{4}} )}{m}, \\
        \mathbf{A_p}^{[2,1]} &= \frac{\minus k_T(\omega^2_{h_{1}} + \omega^2_{h_{2}} + \omega^2_{h_{3}} + \omega^2_{h_{4}} )}{m}, \\
        \mathbf{A_p}^{[4,2]} &= \frac{-b\,(\omega^2_{h_{1}}-\omega^2_{h_{2}}+\omega^2_{h_{3}}-\omega^2_{h_{4}})}{I_{zz}}, \\
        \mathbf{A_p}^{[5,1]} &= \frac{(I_{yy}\,b\,-I_{zz}\,b\,)(\omega^2_{h_{1}}-\omega^2_{h_{2}}+\omega^2_{h_{3}}-\omega^2_{h_{4}})}{I_{yy}\,I_{zz}}, \\
        \mathbf{A_p}^{[6,1]} &= \frac{(I_{yy}\,k_T\,l\, - I_{zz}\,k_T\,l\,)(\omega^2_{h_{1}}-\omega^2_{h_{2}}-\omega^2_{h_{3}}+\omega^2_{h_{4}})}{I_{yy}\,I_{zz}}, \\
        \mathbf{A_p}^{[6,2]} &= \frac{-I_{yy}\,k_T\,l\,(\omega^2_{h_{1}}+\omega^2_{h_{2}}-\omega^2_{h_{3}}-\omega^2_{h_{4}})}{I_{yy}\,I_{zz}}, 
    \end{align}
    }
    where $b$ is the drag constant of the motor with a propeller.
    Matrix $\mathbf{B_p}$ is defined as
    \begin{align}
        \mathbf{B_p} =
        \begin{bmatrix}[1.9]
            \dfrac{k_T}{m} & \dfrac{k_T}{m} & \dfrac{k_T}{m} & \dfrac{k_T}{m}  \\ 
            \dfrac{\minus k_T\,l}{I_{xx}} & \dfrac{\minus k_T\,l}{I_{xx}} & \dfrac{k_T\,l}{I_{xx}} & \dfrac{k_T\,l}{I_{xx}}  \\
            \dfrac{\minus k_T\,l}{I_{yy}} & \dfrac{k_T\,l}{I_{yy}} & \dfrac{k_T\,l}{I_{yy}} & \dfrac{\minus k_T\,l}{I_{yy}}  \\
            \dfrac{\minus b}{I_{zz}} & \dfrac{b}{I_{zz}} & \dfrac{\minus b}{I_{zz}} & \dfrac{b}{I_{zz}} 
        \end{bmatrix}. 
    \end{align}
    We assume a symmetric X-shape of the \ac{UAV} with arm length $l$. 
    Variable $\omega_{h_{i}}$ is the rotor's angular velocity in hover, which is equal to 
    \begin{align}
        k_T \sum_{i=1}^{4} \omega_{h_{i}}^2 = mg.
    \end{align}
    Thanks to the linearized model of the \ac{UAV}, it is possible to utilize linear \ac{MPC} in the trajectory generation process, which is challenging due to the limitations of onboard computational power for searching the large solution space.

\section{MPC-BASED LANDING TRAJECTORY GENERATION}
    In this section, we introduce the \ac{MPC}-based trajectory generator designed specifically for landing a \ac{UAV} on the tilting deck of a moving \ac{USV}.
    Firstly, the derivation of the used \ac{MPC} is detailed in \autoref{sec:model_predictive_control}. 
    It is followed by \autoref{sec:mission_and_navigation} where the state machine responsible for determining the distinct flight mission phases, in which the \ac{UAV} operates, is introduced.
    These stages of flight are used for \ac{MPC}-based trajectory generator, which ensures that the generated trajectory will correspond to the specified flight stage e.g. approaching or landing maneuver.
    This assumption enables dynamically adjusting the defined problem during the process of trajectory generation both in terms of modifying \ac{MPC}'s objective and or its reference.
    Such a process is explained in \autoref{sec:trajectory_generation}.

\subsection{Model Predictive Control}\label{sec:model_predictive_control}
    In general (see \cite{845037}), the objective function of the \ac{MPC} in the tracking form is defined as
    \begin{align}\label{eq:min_opt_mpc}
    \begin{split}
       \min_{\mathbf{e},~\mathbf{u}} ~ &J(\mathbf{e}, \mathbf{u}) = \dfrac{1}{2}\mathbf{e}^\intercal_{[t+N]}\mathbf{P}\mathbf{e}_{[t+N]} + \\
       &\dfrac{1}{2}\sum_{k = 1}^{N-1}\left(\mathbf{e}_{[t+k]}^{\intercal}\mathbf{Q}\mathbf{e}_{[t+k]} + \mathbf{u}^\intercal_{[t+k]}\mathbf{R}\mathbf{u}_{[t+k]}\right),
    \end{split}
    \end{align}
    where $k = \{1, \hdots, N-1\}$ and $N \in \mathbb{Z}^+$ is the size of the prediction and control horizon simultaneously, $t$~stands for the current time, and $\mathbf{e}_{[t+k]}$ denotes the error between the \ac{USV}'s states $\mathbf{r}_{[t+k]}$ and the observable \ac{UAV}'s states $\mathbf{y}_{[t+k]}$.
    Both \ac{UAV} and \ac{USV} models were described in \autoref{sec:usv_uav_models}.
    The penalization matrices $\mathbf{Q}$ and $\mathbf{P}$ are diagonal positive semi-definite matrices. 
    Usually, optimization problems are assumed to be time-invariant, eliminating the need for parameter modifications.
    However, in our case, it is highly beneficial to adjust the parameters of the matrix $\mathbf{Q}$ based on the actual state of the flight, and therefore, from now onward we will assume the matrix to be $\mathbf{Q}$ time varying $\mathbf{Q_{[t]}}$.
    For example, if the \ac{UAV} is following the \ac{USV}, the penalization of the \ac{UAV}'s and \ac{USV}'s linear velocity difference is increased. 
    A detailed description is provided in the following sections.
    The weighting matrix $\mathbf{R}$ is a diagonal positive-definite matrix that penalizes input signals $\mathbf{u}$ in every control step, whereby this input corresponds to a vector containing the square of each rotor's individual angular velocities.
    The \ac{UAV}'s model \refeq{eq:uav_linear_model}, initial state, and both state and input constraints are defined as
    \begin{align}
        \mathbf{x}_{[t+k+1]} &= \mathbf{A}\mathbf{x}_{[t+k]} + \mathbf{B}\mathbf{u}_{[t+k]} ~~~~ \forall k \in \{0, \hdots, N-1 \}, \\
        \mathbf{x}_{[t]} &= \text{measured current state},\\
        \mathbf{x}_{\text{min}} \leq ~&\mathbf{x}_{[t+k]} \leq \mathbf{x}_{\text{max}} ~~~~ \forall k 
        \in \{1, \hdots, N\} ,\\
        \mathbf{u}_{\text{min}} \leq ~&\mathbf{u}_{[t+k]} \leq \mathbf{u}_{\text{max}} ~~~~ \forall k 
        \in \{1, \hdots, N\}.
    \end{align}
    Assuming a steady state of the \ac{UAV} in hover and having a tracking error close to zero, it is not possible to have a control input equal to zero.
    The objective function \refeq{eq:min_opt_mpc} tries to optimize the control input to reduce both the tracking error and the input itself during the optimization process, but it is not aware that the input should not be a nonzero value.
    This approach can increase the deviation between the desired and actual state, thus increasing the reference error.
    Therefore, the standard objective function \refeq{eq:min_opt_mpc} is reformulated.
    Instead of using the input signal $\mathbf{u}_{[t+k]}$ itself, its increment is used:
    \begin{align}\label{eq:utoudelta}
        \Delta \mathbf{u}_{[t]} = \mathbf{u}_{[t]} - \mathbf{u}_{[t-1]}.
    \end{align}
    This leads to the reformulation of a linear model, which must be augmented with the new state variable $\mathbf{x}_{u_{[t-1]}} \cong \mathbf{u}_{[t-1]}$ and the control input is exchanged for $\Delta \mathbf{u}_{[t]}$.
    The form of the augmented model is
     \begin{align} \label{eq:augmented_model1}
        \begin{bmatrix}
        \mathbf{x}_{[t+1]} \\
        \mathbf{x}_{u_{[t+1]}}
        \end{bmatrix}
        &=
        \underbrace{\begin{bmatrix}
        \mathbf{A} & \mathbf{B} \\
        \mathbf{0}^{m\times n} & \mathbf{I}^{m\times m}
        \end{bmatrix}}_{\mathbf{A}_a}
        \underbrace{\begin{bmatrix}
        \mathbf{x}_{[t]} \\ \mathbf{x}_{u_{[t]}}
        \end{bmatrix}}_{\mathbf{x}_{a[t]}}
        +
        \underbrace{\begin{bmatrix}
        \mathbf{B}\\\mathbf{I}^{m\times m}
        \end{bmatrix}}_{\mathbf{B}_a}
        \Delta \mathbf{u}_{[t]}, \\ 
        \mathbf{y}_{[t]} &= 
        \underbrace{\begin{bmatrix}
        \mathbf{C} & \mathbf{0}^{p\times m}
        \end{bmatrix}}_{\mathbf{C}_a}
        \begin{bmatrix}
        \mathbf{x}_{[t]} \\ \mathbf{x}_{u_{[t]}}
        \end{bmatrix},\label{eq:augmented_model2}
    \end{align}
    where $\mathbf{x} \in \mathbb{R}^n$ is a state vector containing $n$ states, $\mathbf{u} \in \mathbb{R}^m$ is the control vector with $m$ inputs, and $\mathbf{y} \in \mathbb{R}^p$ is a system output vector.
    It is also assumed that the primary model omits direct feed-through of the input signal.
    
    The objective function in its sequential form, incorporating an augmented model \refeq{eq:augmented_model1} -- \refeq{eq:augmented_model2}, and excluding constant terms (offsets that can be omitted without affecting the generality of the objective function) is derived by substituting  $\mathbf{e}_{[t]}$ with $\mathbf{r}_{[t]} - \mathbf{C}_a\mathbf{x}_{a[t]}$ into \refeq{eq:min_opt_mpc} resulting in
    \begin{align}
    \begin{split}
        \min_{\Delta \mathbf{u}} &J(\Delta \mathbf{u}, \mathbf{x}_{[t]}, \mathbf{Q}_{[t]}) = \dfrac{1}{2}\Delta \mathbf{u}^\intercal (\bar{\mathbf{C}}_a^\intercal\bar{\mathbf{Q}}_{[t]}\bar{\mathbf{C}}_a + \bar{\mathbf{R}})\Delta \mathbf{u} + \\
        &\begin{bmatrix}
            \mathbf{x}^\intercal_{[t]} & \mathbf{u}^\intercal_{[t-1]} & \mathbf{r}^\intercal_{[t]}
        \end{bmatrix} 
        \begin{bmatrix}
        \bar{\mathbf{A}}_a^\intercal\bar{\mathbf{Q}}_{[t]}\bar{\mathbf{C}}_a \\
        -\bar{\mathbf{T}}\bar{\mathbf{C}}_a
        \end{bmatrix}^\intercal \Delta \mathbf{u}.\label{eq:final_obj_function}
        \end{split}
    \end{align}
    Matrix $\bar{\mathbf{C}}_a$ denotes the controllability matrix of the augmented model of the \ac{UAV}.
    Matrix $\bar{\mathbf{A}}_a$ is a column matrix of the $\mathbf{A}_a$ matrices of the length of the prediction horizon. 
    Matrices $\bar{\mathbf{Q}}$, $\bar{\mathbf{T}}$ and $\bar{\mathbf{R}}$ were obtained from a simplified simultaneous form of the objective function \refeq{eq:min_opt_mpc}.
    
    Since the objective function is introduced in sequential form, the constraints for the states are defined as
    \begin{align}
        \bar{\mathbf{C}}_{a}
        \Delta \mathbf{u} \leq
        \mathbf{x}_{\text{max}}
        -
       \bar{\mathbf{A}}_a \mathbf{x}_{a[t]},
    \end{align}
    where $\mathbf{x}_{\text{max}} \in \mathbb{R}^{(n+m) N}$ are limiting values of the \ac{UAV} state.
    This can be done accordingly for the lower bound.
    Also, slew rate constraints are added, which refer to the range of change of an input variable and are defined as
    \begin{align}
        \Delta\mathbf{u}_{[t]} &\leq \Delta\mathbf{u}_{\text{max}}, \\
        -\Delta\mathbf{u}_{[t]} &\leq -\Delta\mathbf{u}_{\text{min}},\label{eq:input_variable}
    \end{align}
    where $\Delta\mathbf{u}_{\text{max}} \in \mathbb{R}^{(m, N)}$ is a vector of the change of angular speed for individual propellers. 
    The low-level formulation \refeq{eq:final_obj_function} allows us to use the whole linear model of the \ac{UAV} and track position, attitude angles, linear velocities, and Euler rates while still being able to compute the problem in real-time.
    However, not all stages of the flight mission necessitate tracking all \ac{USV} states. 
    The forthcoming section introduces the mission and navigation state machine to address this aspect.

\subsection{Mission and navigation}\label{sec:mission_and_navigation}
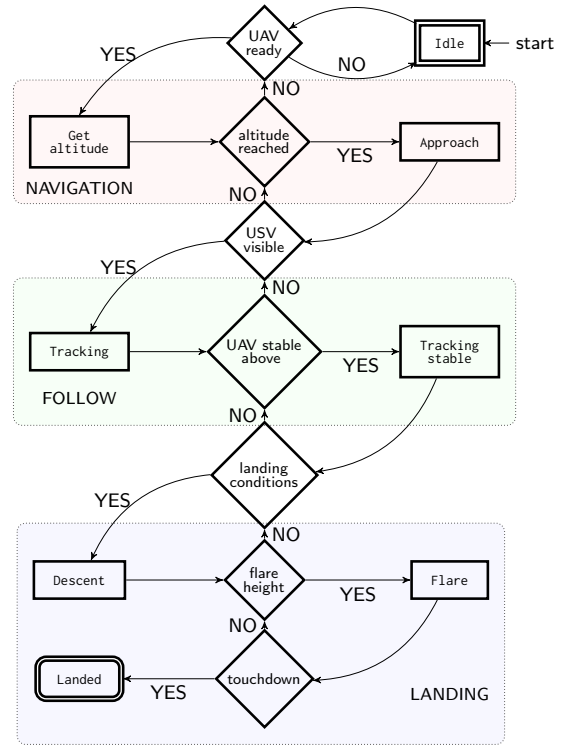
\begin{figure}[t!]
        \centering
        \resizebox{0.45\textwidth}{!}{
            \pgfdeclarelayer{foreground}
\pgfsetlayers{background,main,foreground}
\tikzset{
  state/.style={
    rectangle,
    draw=black, very thick,
    minimum height=1.0em,
    text centered,
  },
  state_diamond/.style={
    diamond,
    draw=black, very thick,
    minimum height=1.0em,
    align=center,
    inner sep=-2ex,
    text centered,
  },
  adder/.style={
    circle,
    inner sep=2pt,
    minimum size=0.3in,
    draw=black, very thick,
    text centered
  },
  state_gray/.style={
    rectangle,
    draw=black, very thick,
    fill=gray!40,
    minimum height=1.0em,
    text centered,
    inner sep=0,
  },
  state_white/.style={
    rectangle,
    draw=black, very thick,
    fill=white,
    minimum height=1.0em,
    text centered,
    text=black,
    inner sep=0,
  },
  state_green/.style={
    rectangle,
    draw=black, very thick,
    fill=green!50,
    minimum height=1.0em,
    text centered,
    text=black,
    inner sep=0,
  },
  state_red/.style={
    rectangle,
    draw=black, very thick,
    fill=red!70,
    minimum height=1.0em,
    text centered,
    text=black,
    inner sep=0,
  },
  state_blue/.style={
    rectangle,
    draw=black, very thick,
    fill=blue!40,
    minimum height=1.0em,
    text centered,
    text=black,
    inner sep=0,
  },
  final_state/.style={
    rectangle,
    rounded corners,
    draw=black, very thick,
    minimum height=2em,
    text centered,
  },
  initial_state/.style={
    rectangle,
    double=white,
    double distance=1pt,
    inner sep=2pt,
    draw=black, very thick,
    minimum height=2em,
    text centered,
  },
  point/.style={
    circle,
    inner sep=0pt,
    minimum size=3pt,
    fill=red
  },
  state_circle/.style={
    circle,
    inner sep=0pt,
    draw=black, very thick,
    minimum size=2em,
    text centered,
  },
  radiation/.style={
  {decorate,
  decoration={
  expanding waves,angle=90,segment length=4pt}
  }},
}

\begin{tikzpicture}[->,>=stealth', node distance=3.0cm,scale=1.0, every node/.style={scale=1.0}]

  
    \node[state, shift = {(0.0, 0.0)}] (get_height) {
      \begin{tabular}{c}
        \footnotesize \texttt{Get} \\[-0.5em]
        \footnotesize \texttt{altitude}
      \end{tabular}
    };

    \node[state_diamond, right of = get_height, shift = {(0.0, 0)}] (if_aproach_altitude) {
      \begin{tabular}{c}
        \scriptsize ~~ altitude ~~\\[-0.5em]
        \scriptsize ~~~ reached ~~
      \end{tabular}
    };
    
    \node[state, right of = if_aproach_altitude, shift = {(0.0, 0.0)}] (approach) {
      \begin{tabular}{c}
        \footnotesize \texttt{Approach}
      \end{tabular}
    };

    \node[state_diamond, below of = if_aproach_altitude, shift = {(.0, 1.4)}] (if_trackin_distance) {
      \begin{tabular}{c}
        \scriptsize ~~ USV ~~ \\[-0.5em]
        \scriptsize ~~~ visible ~~
      \end{tabular}
    };

    \node[state_diamond, below of = if_trackin_distance, shift = {(.0, 1.2)}] (if_uav_above_landing_pattern) {
      \begin{tabular}{c}
        \scriptsize ~~~ UAV stable ~~ \\[-0.5em]
        \scriptsize ~~ above ~~
      \end{tabular}
    };

    \node[state, left of = if_uav_above_landing_pattern, shift = {(0.0, 0.)}] (tracking) {
      \begin{tabular}{c}
        \footnotesize \texttt{Tracking}
      \end{tabular}
    };

    \node[state_diamond, above of = if_aproach_altitude, shift = {(0.0, -1.4)}] (if_conditions_for_planner) {
      \begin{tabular}{c}
        \scriptsize ~~~ UAV ~~ \\[-0.5em]
        \scriptsize ~~~ ready ~~
      \end{tabular}
    };

    \node[initial_state, right of = if_conditions_for_planner, initial, initial where=right, shift = {(0.0, 0.0)}] (idle) {
      \begin{tabular}{c}
        \footnotesize \texttt{Idle} \\
      \end{tabular}
    };
    
    \node[state, right of = if_uav_above_landing_pattern, shift = {(0.0, 0.)}] (tracking_stable) {
      \begin{tabular}{c}
        \footnotesize \texttt{Tracking} \\[-0.5em]
        \footnotesize \texttt{stable}
      \end{tabular}
    };

    \node[state_diamond, below of = if_uav_above_landing_pattern, shift = {(., 1.)}] (if_conditions_for_landing) {
      \begin{tabular}{c}
        \scriptsize ~~~ landing ~~ \\[-0.5em]
        \scriptsize ~~~ conditions ~~
      \end{tabular}
    };

    \node[state_diamond, below of = if_conditions_for_landing, shift = {(.0, 1.3)}] (if_height_for_flare) {
      \begin{tabular}{c}
        \scriptsize ~~~ flare ~~ \\[-0.5em]
        \scriptsize ~~~ height ~~
      \end{tabular}
    };

    \node[state, left of = if_height_for_flare, shift = {(0.0, 0.)}] (descent) {
      \begin{tabular}{c}
        \footnotesize \texttt{Descent}
      \end{tabular}
    };

    \node[state, right of = if_height_for_flare, shift = {(0.0, 0.)}] (flare) {
      \begin{tabular}{c}
        \footnotesize \texttt{Flare}
      \end{tabular}
    };

    \node[state_diamond, below of = if_height_for_flare, shift = {(.0, 1.4)}] (touch_down) {
      \begin{tabular}{c}
        \scriptsize ~~~ touchdown ~~
      \end{tabular}
    };

    \node[final_state,accepting, left of = touch_down, shift = {(0.0, 0.)}] (landed) {
      \begin{tabular}{c}
        \footnotesize \texttt{Landed}
      \end{tabular}
    };
    

    \draw 
        (get_height) edge[below] (if_aproach_altitude)
        (if_conditions_for_planner) edge[left, bend right] node{YES}(get_height.north)
        (idle) edge[below, bend right] (if_conditions_for_planner)
        (if_conditions_for_planner) edge[above, bend right] node{NO} (idle)
        (if_aproach_altitude) edge[] node[below, shift = {(0., -0.00)}]{YES} (approach)
        (if_aproach_altitude) edge[right] node{NO} (if_conditions_for_planner)
        (if_trackin_distance) edge[left] node{NO} (if_aproach_altitude)
        (if_uav_above_landing_pattern) edge[right] node{NO} (if_trackin_distance)
        (approach) edge[below, bend left] (if_trackin_distance)
        (if_trackin_distance) edge[left, bend right] node{YES} (tracking)
        (tracking) edge[below] (if_uav_above_landing_pattern)
        (if_uav_above_landing_pattern) edge[below] node{YES} (tracking_stable)
        (if_conditions_for_landing) edge[left] node{NO}(if_uav_above_landing_pattern)
        (tracking_stable) edge[below, bend left] (if_conditions_for_landing)
        (if_conditions_for_landing) edge[left, bend right] node{YES} (descent)
        (descent) edge[below] (if_height_for_flare)
        (if_height_for_flare) edge[right] node{NO} (if_conditions_for_landing)
        (if_height_for_flare) edge[below] node[shift = {(-0.0, -0.0)}]{YES} (flare)
        (flare) edge[below, bend left] (touch_down)
        (touch_down) edge[below] node[shift = {(-0.0, 0.0)}]{YES} (landed)
        (touch_down) edge[left] node{NO} (if_height_for_flare)
    ;


      \begin{pgfonlayer}{background}
        \path (get_height.west |- if_aproach_altitude.north)+(-0.25,0.25) node (a) {};
        \path (if_aproach_altitude.south -| approach.east)+(+0.25,-0.25) node (b) {};
        \path[fill=red!3,rounded corners, draw=black!70, densely dotted]
          (a) rectangle (b);
      \end{pgfonlayer}
      \node [rectangle, below of=get_height, shift={(0,-0.2)}, node distance=1.7em] (autopilot) {\footnotesize NAVIGATION};

      \begin{pgfonlayer}{background}
        \path (tracking.west |- if_uav_above_landing_pattern.north)+(-0.25,0.25) node (a) {};
        \path (if_uav_above_landing_pattern.south -| tracking_stable.east)+(+0.25,-0.25) node (b) {};
        \path[fill=green!3,rounded corners, draw=black!70, densely dotted]
          (a) rectangle (b);
      \end{pgfonlayer}
      \node [rectangle, below of=tracking, shift={(0, -0.2)}, node distance=1.7em] (autopilot) {\footnotesize FOLLOW};
      
      \begin{pgfonlayer}{background}
        \path (descent.west |- if_height_for_flare.north)+(-0.25,0.25) node (a) {};
        \path (touch_down.south -| flare.east)+(+0.25,-0.25) node (b) {};
        \path[fill=blue!3,rounded corners, draw=black!70, densely dotted]
          (a) rectangle (b);
      \end{pgfonlayer}
      \node [rectangle, below of=flare, shift={(0,-.)}, node distance=5.7em] (autopilot) {\footnotesize LANDING};


\end{tikzpicture}
        }
        \caption{Illustration of the developed state machine composed of three phases ensures decision-making throughout the whole process.}
        \label{fig:state_machine}
        \vspace{-0.5cm}
    \end{figure}

    The \ac{UAV} flight phases before landing are controlled by the finite state machine that actively changes what are the reference states for the MPC-based trajectory planning.
    The proposed state machine is designed to segment the operational mission, assuming the landing trajectory generator can initiate from any location near the \ac{USV}, into several logical phases of flight. 
    This segmentation aims to ensure safety and streamline the overall mission execution. 
    We consider the \ac{UAV} landing procedure to have three phases: NAVIGATION, FOLLOW, and LANDING.
    Individual phases are further divided into states, which are explained in the following paragraph and shown in \autoref{fig:state_machine}.
    \begin{enumerate}
        \item The initial state is \texttt{Idle}.
        The \ac{UAV} remains in this state until the landing command is initiated.
        If the \ac{UAV} is flying and acquiring the \ac{USV}'s position using wireless communication, the state machine passes to the NAVIGATION phase. 
        \item NAVIGATION phase is composed of \texttt{Get altitude} and \texttt{Approach} states.
        \begin{enumerate}
            \item In \texttt{Get altitude} state the \ac{UAV} ascents or descents to a certain approach altitude.
            \item Then, the state machine transitions into the \texttt{Approach} state where the \ac{UAV}'s objective is to move to the \ac{USV}'s location, which is know thanks to the \ac{GNSS} system and mutual communication.
            The \ac{UAV} operates in this state up to the confirmation of visual contact of the \ac{USV} when the \ac{USV}'s state prediction starts using visual relative localization systems instead of using \ac{GNSS}, which can suffer from drifts. 
            At that moment, the state machine passes to the next phase.
        \end{enumerate}
        \item FOLLOW phase is composed of \texttt{Tracking} and \texttt{Tracking stable} states.
        The goal of this state is to follow the \ac{USV}.
        \begin{enumerate}
            \item The first state of the FOLLOW phase is \texttt{Tracking}, which represents the \ac{UAV}'s final approach above the \ac{USV}'s deck to the tracking altitude.
            \item When the \ac{UAV} is above the deck at a predefined altitude, the state machine passes to \texttt{Tracking stable} and waits for the appropriate conditions for landing while the \ac{UAV} is tracking the \ac{USV}.
            These conditions are the distance threshold from the \ac{USV}'s center in the horizontal plane, the predicted trajectory of the \ac{USV} must be feasible for the \ac{UAV}, and the estimated \ac{USV}'s position covariance must be in a predefined limit.
            When all conditions for landing are met, the state machine passes to the next phase.
       \end{enumerate}
       \item LANDING phase is composed of \texttt{Descent}, \texttt{Flare}, and \texttt{Landed} states.
       \begin{enumerate}
           \item The \texttt{Descent} state represents the initial stage of the landing maneuver in which the \ac{UAV} starts descending towards the \ac{USV}'s deck.
           If one of the landing conditions is violated or the new measurements are not received within a predefined time limit, the state machine transitions to the state \texttt{Tracking}.
           \item The previous state passes to the state \texttt{Flare} when the \ac{UAV} reaches a certain height above the \ac{USV}'s deck. 
           This maneuver is similar to an airplane's final stage of landing approach in which \ac{UAV} tries to match the attitude with the \ac{USV}'s deck in order to land smoothly on all legs.
           The touchdown is verified by monitoring the estimation of the needed thrust to maintain the flight and verified using accelerometers and the range finder sensor.
           In case of missing the landing platform and losing sight of the \ac{USV}'s deck the \ac{UAV} will return to the phase FOLLOW.
           For these emergency situations, the trajectory generator has the minimum altitude set which prevents generating the trajectory into the water.          
           \item After touching down, the \ac{UAV}'s motors are turned off and the state machine passes to the \texttt{Landed} state.
       \end{enumerate}
    \end{enumerate}

    By dividing the landing problem into smaller individual states, it is possible to customize the \ac{MPC}-based trajectory generation and the reference passed to the \ac{MPC} for each state of flight.

\subsection{Trajectory generation}\label{sec:trajectory_generation}
    The \ac{MPC}-based trajectory generation aims to create references for the \ac{UAV} based on the \ac{UAV}'s flight phase.
    In the first phase of the \ac{UAV}'s operation (NAVIGATION), a point reference in the approach altitude $h_a$ and directly above the current \ac{UAV}'s position is used.
    Moreover, the \ac{UAV}'s heading reference is adjusted, such that the camera on the \ac{UAV} aims towards the \ac{USV}'s location.
    The predicted x, y coordinates of the \ac{USV} are used for the \texttt{Approach state}.
    Hence, the trajectory from the \ac{MPC} ensures that the control leads the \ac{UAV}'s position in such a way that the \ac{UAV} performs an approach to the \ac{USV}'s position at a given approach altitude $h_a$ and heading $h$ directing towards the \ac{USV}.
    Due to this, the camera can be mounted on the drone under a certain angle, enabling efficient use of the camera's \ac{FOV}.
    For the FOLLOW phase, the predicted velocity of the \ac{USV} is added to the reference.
    The \ac{UAV} is thus able to follow the moving \ac{USV}, maintaining the predefined altitude $h_t$.

    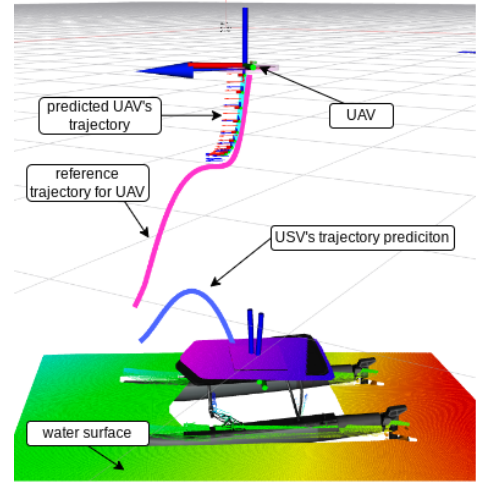
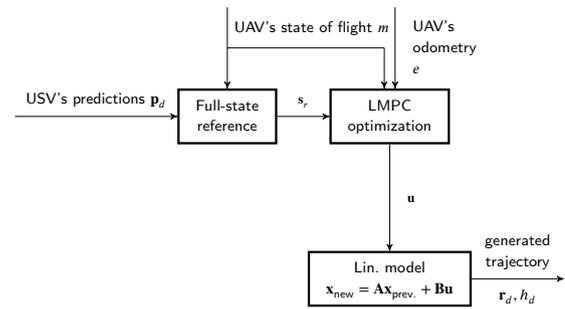
\begin{figure}[t!]
        \centering
        \subfloat[The visualization in Rviz illustrates the creation of the guiding path for the MPC-based trajectory generator, depicting the beginning of the landing maneuver.]{
        \includegraphics[width = 0.35\textwidth]{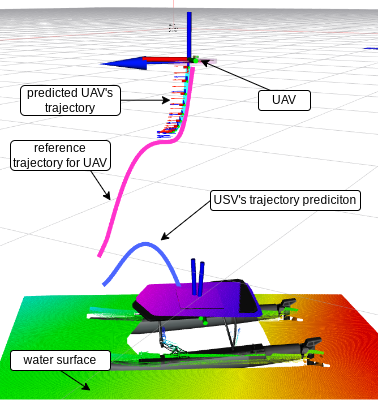}
        \label{fig:uav_descent_maneuver}
        }
        \hspace{0.6cm}
        \subfloat[Diagram of the trajectory generator, where the UAV's flight state comes from the state machine and $\mathbf{s}_r$ is a modified full-state reference of the size of the prediction horizon that the MPC tracks. The MPC's output is a sequence of action interventions $\mathbf{u}$, which are fed one by one into the linear model and from which the desired trajectory ($\mathbf{r}_d$, $h_d$) is created.]{
        \raisebox{0.75cm}%
        {\resizebox{0.45\textwidth}{!}{
            \pgfdeclarelayer{foreground}
\pgfsetlayers{background,main,foreground}
\tikzset{
  state/.style={
    rectangle,
    draw=black, very thick,
    minimum height=1.0em,
    text centered,
  },
  smallstate/.style={
    rectangle,
    draw=black, very thick,
    minimum height=0.2em,
    text centered,
  },
  final_state/.style={
    rectangle,
    rounded corners,
    draw=black, very thick,
    minimum height=2em,
    text centered,
  },
  initial_state/.style={
    rectangle,
    double=white,
    double distance=1pt,
    inner sep=2pt,
    draw=black, very thick,
    minimum height=2em,
    text centered,
  },
  point/.style={
    circle,
    inner sep=0pt,
    minimum size=3pt,
    fill=red
  },
  adder/.style={
    circle,
    inner sep=2pt,
    minimum size=0.3in,
    draw=black, very thick,
    text centered
  },
  state_gray/.style={
    rectangle,
    draw=black, very thick,
    fill=gray!40,
    minimum height=1.0em,
    text centered,
    inner sep=0,
  },
  state_white/.style={
    rectangle,
    draw=black, very thick,
    fill=white,
    minimum height=1.0em,
    text centered,
    text=black,
    inner sep=0,
  },
  state_green/.style={
    rectangle,
    draw=black, very thick,
    fill=green!50,
    minimum height=1.0em,
    text centered,
    text=black,
    inner sep=0,
  },
  state_red/.style={
    rectangle,
    draw=black, very thick,
    fill=red!70,
    minimum height=1.0em,
    text centered,
    text=black,
    inner sep=0,
  },
  state_blue/.style={
    rectangle,
    draw=black, very thick,
    fill=blue!40,
    minimum height=1.0em,
    text centered,
    text=black,
    inner sep=0,
  },
  final_state/.style={
    rectangle,
    rounded corners,
    draw=black, very thick,
    minimum height=2em,
    text centered,
  },
  initial_state/.style={
    rectangle,
    double=white,
    double distance=1pt,
    inner sep=2pt,
    draw=black, very thick,
    minimum height=2em,
    text centered,
  },
  point/.style={
    circle,
    inner sep=0pt,
    minimum size=3pt,
    fill=red
  },
  radiation/.style={
  {decorate,
  decoration={
  expanding waves,angle=90,segment length=4pt}
  }},
}

\begin{tikzpicture}[->,>=stealth', node distance=3.0cm,scale=1.0, every node/.style={scale=1.0}]

    \node[state, shift = {(0.0, 0.0)}] (ref_predict) {
      \begin{tabular}{c}
        \footnotesize Full-state \\
        \footnotesize reference
      \end{tabular}
    };
    
  \node[state, right of = ref_predict , shift = {(0.0, 0.0)}] (controller) {
      \begin{tabular}{c}
        \footnotesize LMPC \\
        \footnotesize optimization
      \end{tabular}
    };

    \node[state, below of = controller, shift = {(0, 0)}] (model) {
      \begin{tabular}{c}
        \footnotesize Lin. model \\
        \footnotesize $\mathbf{x}_{\text{new}} = \mathbf{A}\mathbf{x}_{\text{prev.}}+\mathbf{Bu}$
      \end{tabular}
    };
    

    \path[->] ($(ref_predict.north) + (0.0, 1.5)$) edge [] node[right, near start, shift = {(-0.20, 0.0)}] {
      \begin{tabular}{l}
        \footnotesize UAV's state of flight $m$
    \end{tabular}}($(ref_predict.north) + (0.0, 0)$);
    
    \draw[-] ($(ref_predict.north)+(0, 0.75)$) edge ($(ref_predict.north |- ref_predict.north)+(0.0,0.75)$) -- ($(controller.north |- controller.north)+(-0.1,0.75)$) edge [->,] ($(controller.north)+(-0.1, 0)$);
    
    \path[->] ($(ref_predict.west) + (-3, -0.)$) edge [] node[above, midway, shift = {(0.0, 0.0)}] {
      \begin{tabular}{c}
        \footnotesize USV's predictions $\mathbf{p}_d$
    \end{tabular}}($(ref_predict.west) + (0.0, -0.)$);
    
    \path[->] ($(controller.north) + (0.1, 1.5)$) edge [] node[right, midway, shift = {(0.0, 0.0)}] {
      \begin{tabular}{l}
        \footnotesize UAV's \\
        \footnotesize odometry \\
        \footnotesize $e$
    \end{tabular}}($(controller.north) + (0.1, 0)$);
    
    \path[->] ($(ref_predict.east) + (0.0, 0.0)$) edge [] node[above, midway, shift = {(0.0, 0.00)}] {
      \begin{tabular}{c}
        \scriptsize $\mathbf{s}_r$
    \end{tabular}} ($(controller.west) + (0.0, 0.00)$);

    \path[->] ($(model.east) + (0.0, 0)$) edge [] node[above, midway, shift = {(0.0, 0.0)}] {
      \begin{tabular}{c}
        \footnotesize generated \\
        \footnotesize trajectory
    \end{tabular}}
    node[below, midway, shift = {(0.0, 0.0)}] {
      \begin{tabular}{c}
        \footnotesize $\mathbf{r}_d, h_d$
    \end{tabular}}($(model.east) + (1.7, 0)$);

    \path[->] ($(controller.south) + (0.0, 0.0)$) edge [] node[right, midway, shift = {(0.0, 0.00)}] {
      \begin{tabular}{c}
        \scriptsize $\mathbf{u}$
    \end{tabular}} ($(model.north) + (0.0, 0.00)$);

\end{tikzpicture}
        }}
        \label{fig:reference_tracker_new}
        }
        \caption{Visualization of the landing maneuver in Rviz together with the trajectory generator pipeline diagram.}
        \vspace{-0.3cm}
    \end{figure}
    During a descent maneuver, the reference is a trajectory generated by sampling the z-position to maintain a specific vertical approach velocity relative to the \ac{USV}'s deck.
    Additionally, a vertical velocity reference is established to maintain a constant mutual vertical speed.
    The visualization of such obtained guided path can be seen in \autoref{fig:uav_descent_maneuver}.
    Throughout the flare maneuver, the predicted position, linear velocity, attitude angles, and Euler rates of the \ac{USV}'s deck are supplied to the \emph{MPC optimization} (see \autoref{fig:reference_tracker_new}) to create a trajectory that minimizes both position and attitude deviation upon the touchdown.
    Thanks to the continuous creation of the trajectory, the \ac{UAV} can react to the latest \ac{USV} state.
    
    However, even when tracking all \ac{USV} states during the final flare maneuver, it does not guarantee a smooth touchdown regarding attitude alignment.
    It is because of the fact that the \ac{UAV}'s horizontal acceleration is controlled by tilting, and therefore, change in roll and pitch angle influences the velocity and the position of the \ac{UAV}, which is also demanded to be tracked as precisely as possible.
    To address this challenge, we propose a method that dynamically changes penalization matrix $\mathbf{Q}_{[t]}$ from \refeq{eq:final_obj_function} to get the most precise tracking of the certain states such as attitude angles upon the touchdown.
    For this purpose, we introduce an exponential function
    \begin{align}\label{eq:q_change}
        f(v_d) = 1 + \dfrac{\alpha}{e^{\beta v_d}}
    \end{align}
    that is used to change the elements of the penalization matrix $\mathbf{Q}_{[t]}$, specifically roll and pitch angles.
    This function depends on the vertical distance $v_d$ between the \ac{UAV} and the \ac{USV}'s deck.
    The coefficients $\alpha$ and $\beta$ are tuned to increase the value of the specific elements of the $\mathbf{Q}_{[t]}$ in sufficient distance to have time for the attitude synchronization maneuver, which we call \ac{FAA}.
    This can be done only when the approach velocity is constant, which was already achieved by adjusting the reference vertical speed, as mentioned above.
    As the \ac{UAV} approaches the deck, the $\mathbf{Q}_{[t]}$'s matrix weights belonging to the attitude angles exponentially increase, resulting in prioritizing the correction of attitude deviations over other tracked \ac{USV}'s states.
    It is also noteworthy that aligning the attitude is advantageous even when landing on a rapidly moving \ac{USV}.

\begin{figure*}[!t]
        \centering
        \resizebox{1.0\textwidth}{!}{
            \input{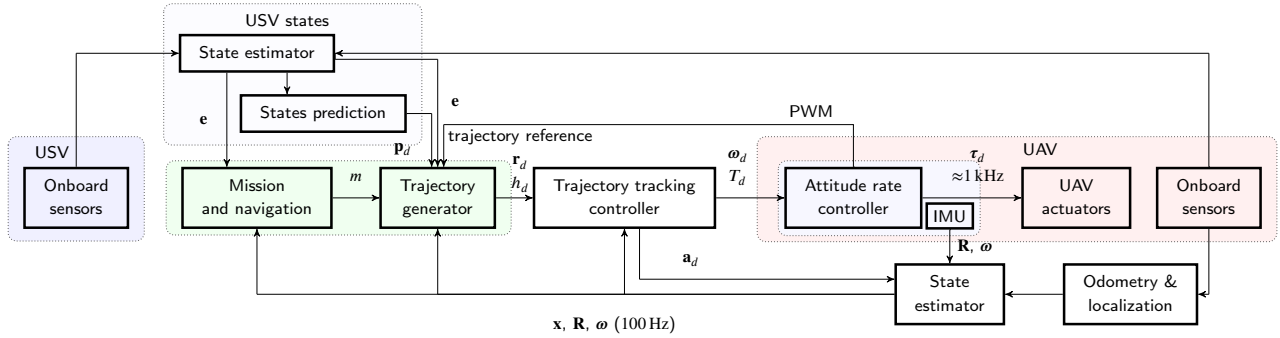}
        }
        \caption{Full system pipeline diagram including USV's estimation/prediction, mission planning, trajectory planning, and the Multi-robot Systems multirotor control system \protect\cite{baca2021mrs}. 
        \emph{Mission and navigation} software supplies the flight mode ($m$) to the \emph{Trajectory generator} where the reference trajectory composed of headings and positions ($\mathbf{r}_d$, $h_d$) is generated. 
        \emph{Trajectory tracking controller} ensures the trajectory is precisely tracked by producing the desired thrust and angular velocities ($T_d$, $\bm{\omega}_d$) for the Pixhawk embedded flight controller. 
        The \emph{State estimator} fuses data from \emph{Onboard sensors} and \emph{Odometry \& localization} methods to create an estimate of the UAV translation and rotation ($\mathbf{x}$, $\mathbf{R}$).
        Estimated states ($\mathbf{e}$) and a full-states prediction ($\mathbf{p}_d$) of the USV are provided by the \emph{State estimator} and \emph{States prediction} software, which exploits the \emph{Onboard sensors} of the UAV and USV.}
        \label{fig:pipeline_diagram}
        \vspace{-0.7cm}
    \end{figure*}

    In the final stage of the \ac{MPC}-based trajectory generator the \ac{MPC}'s output is not used directly, instead the trajectory is computed using angular rates of the individual motor obtained from \ac{MPC} and \ac{UAV}'s linear model.
    The generated trajectory serves as the tracking trajectory for the \textit{Reference tracker} in \ac{MRS} system \cite{baca2018model}.
    A graphical representation of the trajectory generator interconnection into the \ac{MRS} \ac{UAV} control system is shown in \autoref{fig:pipeline_diagram}.

\section{SIMULATION VERIFICATION}
    The integration of the proposed method into the \ac{MRS} system \cite{baca2021mrs, hert2023mrs} enabled realistic verification in a simulated environment, which allowed us to test the system under various conditions and even in high wind.
    The \ac{MRS} system estimates the wind-based forces and deflections, which are then corrected by the attitude rate controller.
    As shown in \reftab{tab:wave_definition}, the most probable waves have a height between \SI{1.25}{\meter} and \SI{2.5}{\meter}. 
    Therefore, we primarily focused on designing the overall system using the proposed method to efficiently land the \ac{UAV} on the \ac{USV} located in \textit{Moderate sea}.
    In the first simulation experiment, the \ac{USV} is left uncontrolled, drifting with the water current in \autoref{sec:carried_by_current}. 
    Then, the \ac{USV} is following a predefined path in \autoref{sec:usv_fol_pre_path}.
    Lastly, the proposed approach was compared to the most related state-of-the-art approach \cite{gupta2022landing} in \autoref{state_of_the_art_comparison}.

\subsection{Simulation Environment}\label{sec:sim_environment}
\begin{figure*}[t!]
        \centering
        \subfloat[An image from the down-facing camera mounted on the UAV.]{\fbox{\includegraphics[width=0.319\textwidth, angle=0]{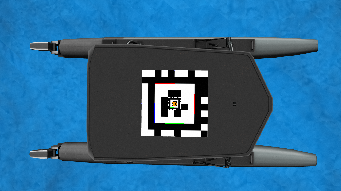}}}
        \hspace{0.1cm}
        \subfloat[Screenshot from the Gazebo simulator depicting a UAV above the USV's deck during the landing maneuver.]{\fbox{\includegraphics[width=0.31\textwidth]{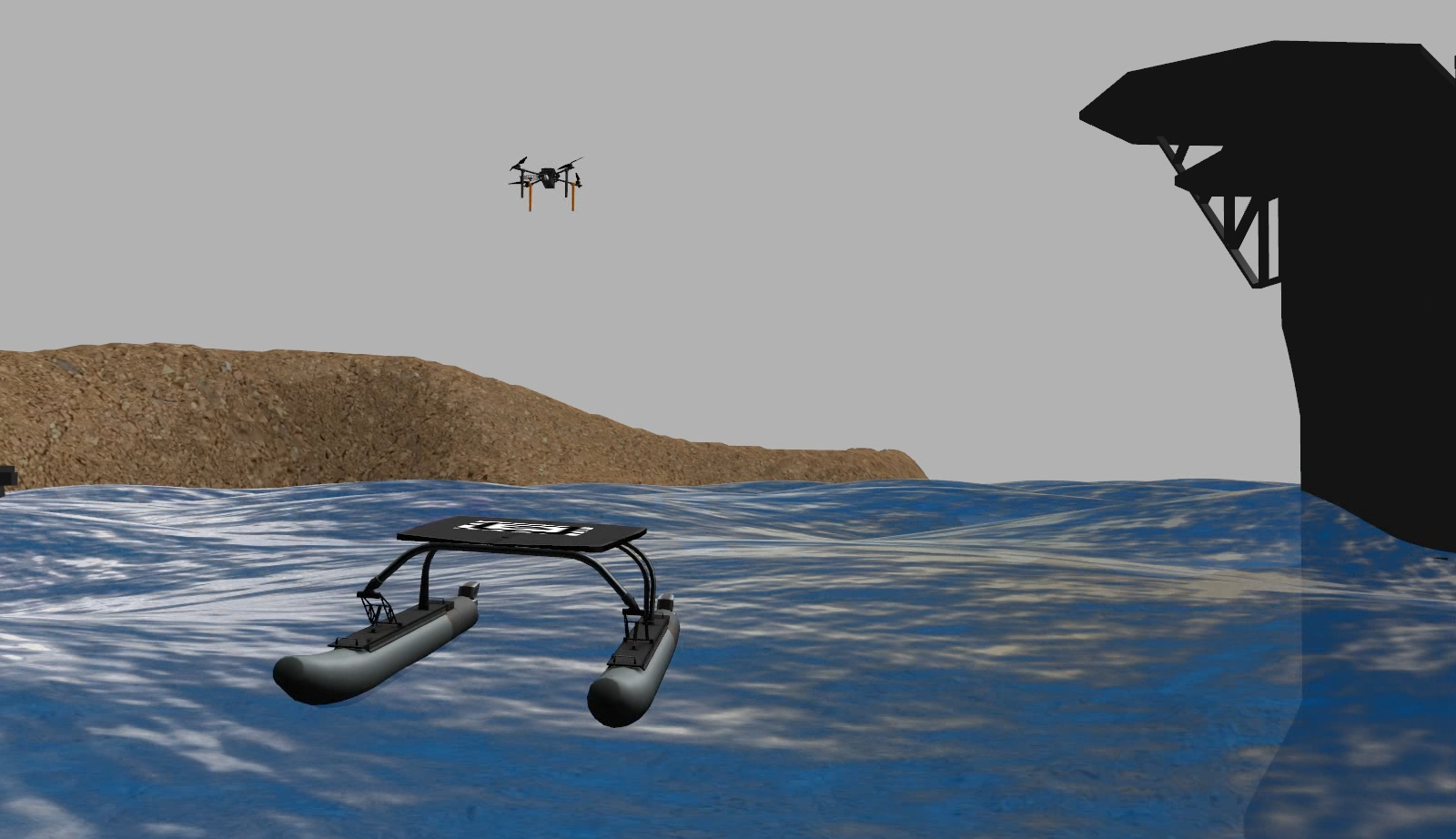}}} 
        \hspace{0.1cm}%
        \subfloat[Detail of the landed UAV on the USV's deck.]{ 
        {\fbox{\includegraphics[clip, trim={0 60 0 60}, width=0.3\textwidth]{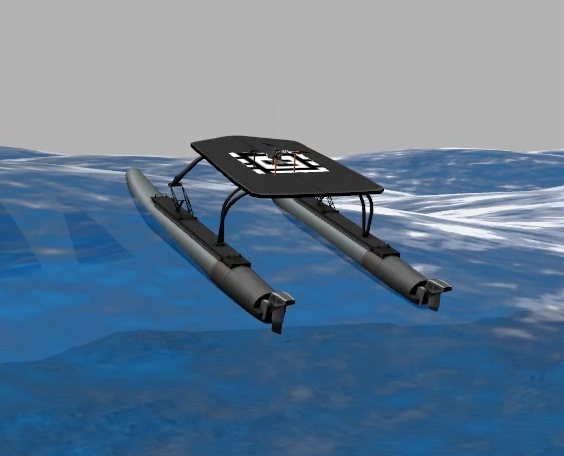}}}}
        \caption{Screenshots from the Gazebo simulator during the UAV's landing maneuver on \textit{Moderate sea}.
        }
        \label{fig:gazebo_rviz_overview}
        \vspace{-0.5cm}
    \end{figure*}

    The proposed approach was thoroughly tested in a simulation with highly realistic waves that was inspired by the project called \ac{VRX} \cite{bingham19toward}, which was created for the maritime \ac{VRX} Competition.
    The designation of the waves together with their heights is denoted in \reftab{tab:wave_definition}.
    It is also worth mentioning that the height of the waves also correlates with wind speed, as shown in \cite[p.~13]{abujoub2019development}. 
    However, the wind is omitted in the simulations.
    The simulation environment based on the \ac{MRS} Gazebo simulation \footnote{\url{https://github.com/ctu-mrs/mrs_uav_gazebo_simulation}} was created (see \autoref{fig:gazebo_rviz_overview}).
    The \ac{UAV}, which was used in simulations, weights \SI{3.5}{\kilo\gram}, is \SI{0.15}{\meter} in height and the arm length is \SI{0.325}{\meter}.
    The \ac{USV} is \SI{5}{\meter} long, \SI{2.5}{\meter} in width, \SI{1.3}{\meter} in height and weights around \SI{180}{\kilo\gram}.
    Its landing platform, which measures \SI{2.5}{\meter} in length and \SI{1.7}{\meter} in width, is covered with the AprilTag with a side length of \SI{1}{\meter}.
    The \ac{USV} is equipped with an engine mounted on a joint at the back of each float, allowing it to navigate through predefined waypoints using a simple \ac{PID} controller.
    \begin{table}[t!]
        \centering
        \caption{Definition of sea state codes \protect\cite{bishop1974modal}. Notice that the percentage probability for Calm and Smooth seas are summed up.
        The table is a modified version of the one from \cite[p.~200]{fossen2011handbook}.}
        \label{tab:wave_definition}
        \begin{threeparttable}
        \begin{tabular}{lcccc}
        \hline
                          &    \SI{}{\meter}      & \multicolumn{3}{c}{Percentage probability \si{\percent}}                                   \\ \hline
        \begin{tabular}[c]{@{}c@{}}Description \\ of sea\end{tabular} &
          \begin{tabular}[c]{@{}c@{}}Wave \\ height \\ observed \end{tabular} &
          \begin{tabular}[c]{@{}c@{}}World\\ wide\end{tabular} &
          \begin{tabular}[c]{@{}c@{}}North\\ Atlantic\end{tabular} &
          \begin{tabular}[c]{@{}c@{}}Northern \\ North \\ Atlantic\end{tabular} \\ \hline
        Calm\tnote{1}     & 0        & \multirow{3}{*}{11.2486} & \multirow{3}{*}{8.3103} & \multirow{3}{*}{6.0616} \\
        Calm\tnote{2}    & 0--0.1    &                          &                         &                         \\
        Smooth\tnote{3} & 0.1--0.5  &                          &                         &                         \\
        Slight            & 0.5--1.25 & 31.6851                  & 28.1996                 & 21.5683                 \\
        Moderate          & 1.25--2.5 & 40.1944                    & 42.0273                 & 40.9915                 \\
        Rough             & 2.5--4.0  & 12.8005                  & 15.4435                 & 21.2383                 \\
        Very rough        & 4.0--6.0  & 3.0253                   & 4.2938                  & 7.0101                  \\
        High              & 6.0--9.0  &  0.9263                  &  1.4968                 & 2.6931                  \\
        Very high         & 9.0--14.0 & 0.1190                   &  0.2263                  & 0.4346                 \\
        Phenomenal        &Over 14.0 & 0.0009                  &  0.0016                &    0.0035               \\ \hline
        \end{tabular}
        \begin{tablenotes}
            \item[-] (1) glassy, (2) rippled, (3) wavelets
        \end{tablenotes}
        \end{threeparttable}
        \vspace{-0.4cm}
    \end{table}
    
     
    Water motion is modeled using the summation of Gerstner waves that represent the water surface as a trochoidal shape~\cite{tessendorf2001simulating}. 
    This allows to adjust the amplitude, period, direction, and steepness of the waves.
    The horizontal and vertical displacements of the wave-field to the undisturbed horizontal location $\mathbf{x}_0 = [x_0, y_0]^\top$ with a vertical height of $\zeta_0 = 0$ are defined as
    \begin{align}\label{eq:gerstner_waves}
        \mathbf{x}(\mathbf{x}_0, t) &= \mathbf{x}_0 - \sum_{i=1}^{N_{w}} q_i(\bm{v}_i/v_i) A_i \sin{(\bm{v}_i\mathbf{x}_0 - \omega_{w_{i}} t + \phi_i)},\\
        \zeta(\mathbf{x}_0, t) &= \sum_{i=1}^N A_i \cos{(\bm{v}_i\mathbf{x}_0 - \omega_{w_{i}} t + \phi_i)}.
    \end{align}
    Each wave component is defined with steepness $q_i$, amplitude $A_i$, angular frequency $\omega_{w_{i}}$, random phase $\phi_i$, and wave vector $\bm{v}_i$ in the direction of travel with the magnitude of $v_i$.
    The $N_w$ in \refeq{eq:gerstner_waves} is the number of wave components at time $t$.
    In our case, three wave components $N_w$ were used to simulate an offshore environment.
    It is also worth mentioning that waves with a steepness greater than 1:7 (wave height to wavelength) are considered breaking waves, which means that they are breaking at their crest \cite{https://doi.org/10.1029/2009GL041771}.
    Therefore, the nonbreaking waves can have a maximum inclination of around \SI{25}{\degree}.
    We assume that the \ac{UAV} together with the \ac{USV} will not be operated in worse conditions for safety reasons.
    This implies that the \ac{UAV}'s landing maneuver won't be performed in such conditions. 

\subsection{USV state estimation and prediction}
    The core component of the state estimation and prediction system is the Kalman filter, which fuses data from multiple sensors to increase the method's reliability.
    The system uses a mathematical model of the \ac{USV} in the prediction step, which can provide temporary estimates even if no new sensory data are received for a limited time.
    The accuracy of predicted states depends on the accuracy of the \ac{USV} model (see \autoref{sec:usv_uav_models}) and the correctness of its defining parameters, which specify the \ac{USV}'s properties.
    The potential imprecision of the model identification impacts the prediction of the \ac{USV} states used by the trajectory generator, which can lead to an imprecise landing.
    The predicted \ac{USV} states are position $[\hat{b}_x, \hat{b}_y, \hat{b}_z]^\top$, attitude $[\phi_b, \theta_b, \psi_b]^\top$, linear velocity $[u_b, v_b, w_b]^\top$, and Euler rates $[p_b, q_b, r_b]^\top$.
    
    The state estimation system relies on mutual communication, sharing the \ac{GNSS} location and \ac{IMU} data.
    For precise localization during the FOLLOW and LANDING phases, visual relative localization is employed using the AprilTag detector and \ac{UVDAR} system, and even the \ac{GNSS} information can be utilized in extreme cases in these operational phases.
    Additionally, an \ac{IMU} is employed for the attitude and angular velocity estimation.
    This phase heavily relies on the precision and noise levels of sensory measurements.
    Extensive verification of the used sensors, as detailed in \cite{tro2023novak}, indicates that the AprilTag detector delivers the lowest error in position and linear velocity estimation, whereas the \ac{IMU} excels in attitude and Euler rates estimation.
    According to \cite{tro2023novak}, the estimation performance measured with \ac{RMSE} under \textit{Moderate sea} conditions of the used estimation system is following.
    Position \ac{RMSE} is equal to \SI{0.116}{\meter}, attitude \ac{RMSE} is 
    \SI{0.017}{\radian}
    , linear velocity \ac{RMSE} is \SI{0.201}{\meter\per\second} and angular velocity \ac{RMSE} is 
    \SI{0.004}{\radian\per\second}
    .

\subsection{MPC-based trajectory generation settings}\label{sec:ref_lim_set}
    The configuration of the \ac{MPC}-based trajectory generator is based on the attributes of the employed \ac{UAV} and our practical experiences.
    However, this does not limit our method to a specific type of \ac{UAV}, as the \ac{UAV} model presented in \autoref{sec:usv_uav_models} was described in a general manner and therefore the mathematical model can be tailored to a specific \ac{UAV}.

     We summarize the most important parameters of our method, such as the altitude for approach state or vertical velocity in landing state in \autoref{table:traj_set}.
    These parameters were obtained empirically and can be changed dynamically at any time during the flight.
    \begin{table}[t!]
        \centering
        \caption{Trajectory generator settings and limits for trajectory state machine.\\\hspace{\textwidth}}\label{tab:simulation_settings_all}
               \label{table:traj_set}  
            \vspace{-0.4cm}
            \begin{tabular}{lll}
            \hline
            Symbol  & Value                  & Description          \\ \hline \hline
            $h_{a}$ & $\SI{15}{\meter}$      & approach altitude \\
            $h_{t}$ & $\SI{7}{\meter}$      & tracking altitude \\
            $v_{la}$ & $\SI{1}{\meter\per\second}$      & landing approach velocity \\
            $v_{fa}$ & $\SI{0.5}{\meter\per\second}$    & flare approach velocity \\
            \hline
            \end{tabular}
    \end{table}
    We also put constraints on both the linear and angular velocities based on the limitations of the \textit{Reference tracker}, as shown in the block diagram in \autoref{fig:pipeline_diagram}. 
    These limits are listed in \autoref{tab:trajectory_planner_settings}.
    These parameters can be adjusted according to the used \ac{UAV}. 
    However, it must be noted that if the limits are too low, the \ac{UAV} may never perform a landing maneuver in harsh sea conditions. 
    This is because the \ac{USV}'s movements may not be achievable by the \ac{UAV}.

    It was experimentally found that the maximum speed of the \ac{USV} on which is our method capable of successfully landing is \SI{7}{\meter\per\second} in \textit{Calm sea} environment.
    On the other hand, the same speed of the \ac{USV} is not achievable in harsh sea conditions due to the acceleration limitations within which the state estimator is capable of providing an accurate estimation.

    Another important parameter of the proposed method is the length of the generated trajectory.
    In this work, we use \SI{2}{\second} prediction horizon with 20 prediction steps.
    The length of the trajectory was determined based on the computation complexity.
    Our analysis showed that the trajectory generator can operate at a maximum frequency of \SI{10}{\hertz}, allowing sufficient time for the solver to converge in the NAVIGATION phase.
    During the FOLLOW and LANDING phases, the operation rate increases to \SI{50}{\hertz} since the solution is found within a maximum of \SI{20}{\milli\second}.
    The inconsistent computation time is handled by the MRS system trajectory tracker, which performs the interpolation of the given trajectory to find the correct time and state to follow and which ensures smooth tracking of a given trajectory.
    
    \begin{table*}[!b]
        \centering
        \caption{Maximal allowed constraints for UAV model in trajectory generator to generate landing trajectories for the \ac{UAV} and penalization matrices used in the objective function.}
        \label{tab:trajectory_planner_settings}
        \begin{threeparttable}
            \begin{tabular}{lll}
            \hline
            Symbol  & Value                  & Description          \\ \hline \hline
            $\mathbf{R}$      & \tnote{*} $\text{blkdiag}[0.1, 0.1, 0.1, 0.1]$      & control input weighting matrix \\
            $\mathbf{Q}$      & \tnote{*} $\text{blkdiag}[30, 30, 40 + \frac{10000}{e^{20d}}, 1 + \frac{50000}{e^{10d}}, 1 + \frac{50000}{e^{10d}}, 50, 1, 1, 3000 + \frac{3000}{e^{25d}}, 1, 1, 1]$    &  state weighting matrix \\
            $\mathbf{P}$ & \tnote{*} $\text{blkdiag}[0,0,0,0,0,0,0,0,0,0,0,0]$ & final state weighting matrix \\
            $\phi,~\theta$      & $\pm\SI{0.7854}{\radian}$ & roll and pitch min/max angle  \\
            $v_x,~v_y$      & $\pm\SI{8}{\meter\per\second}$      & horizontal velocity in x and y axis \\
            $v_z$      & $\pm\SI{4}{\meter\per\second}$ & vertical min/max velocity  \\
            $v_\phi,~v_\theta,~v_\psi$      & $\pm\SI{2}{\radian\per\second}$      & roll, pitch and yaw rotation velocity \\ \hline
            \end{tabular}
            \begin{tablenotes}
                \item[*] blkdiag denotes block diagonal matrix
            \end{tablenotes}
        \end{threeparttable}
        \vspace{-0.5cm}
    \end{table*}

 \subsection{Landing on a USV carried by current}\label{sec:carried_by_current}
    In the first set of experiments, the \ac{USV} was located on a \textit{Moderate sea} and was passively carried by the water current, which is a typical situation in many \ac{UAV}-\ac{USV} scenarios.
    Empirical findings indicate that setting the Gerstner wave's peak amplitude to \SI{0.5}{\meter} and the period to \SI{5}{\second} resulted in waves with heights ranging from \SI{0.6}{\meter} to \SI{2.8}{\meter} in simulation.
    The position and heading were randomly generated within a predefined range for the experiments.

    \begin{figure}[b!]
        \centering
        \input{trajectory_generation}
        \caption{Comparison of the MPC-based trajectory generator with FAA at the final pre-touchdown maneuver and without FAA in simulation. The vertical yellow line indicates a touchdown.}
        \label{fig:matlab_motivation_landing}
        \vspace{-0.5cm}
    \end{figure}
    The recorded motion of the \ac{USV} was employed to fine-tune the $\alpha$ and $\beta$ coefficients from \refeq{eq:q_change} for \ac{MPC}-based trajectory generator using \ac{FAA}.
    The comparison of \ac{MPC}-based trajectory generator with and without \ac{FAA} is demonstrated in \autoref{fig:matlab_motivation_landing}. 
    The graphs show the \ac{UAV}'s horizontal and vertical position together with attitude-tracking reference and performed tracking of the \ac{UAV} during the simulated landing maneuver.
    The trajectory generated by the \ac{MPC}-based trajectory generator utilizing \ac{FAA} ensured synchronization of attitude angles at the touchdown.
    In contrast, the \ac{MPC}-based trajectory generator without \ac{FAA} does not assure this synchronization.
    If we compare position precision while using \ac{FAA}, the performed position deteriorated only \SI{1}{\centi\meter} in the x-axis and \SI{1.5}{\centi\meter} in the y-axis.    
    This demonstrates that the \ac{FAA} approach ensures better synchronization of the \ac{UAV}'s and \ac{USV}'s attitude angles upon the touchdown without any apparent deterioration in position precision.

    \begin{figure}[b!]
        \centering
        \subfloat{\begin{tikzpicture}[]
            \hspace{-0.3cm}
            \node[anchor=south west,inner sep=0] (a) at (0,0) { \includegraphics[clip, trim={30 226 67 215}, width = 0.5\textwidth]{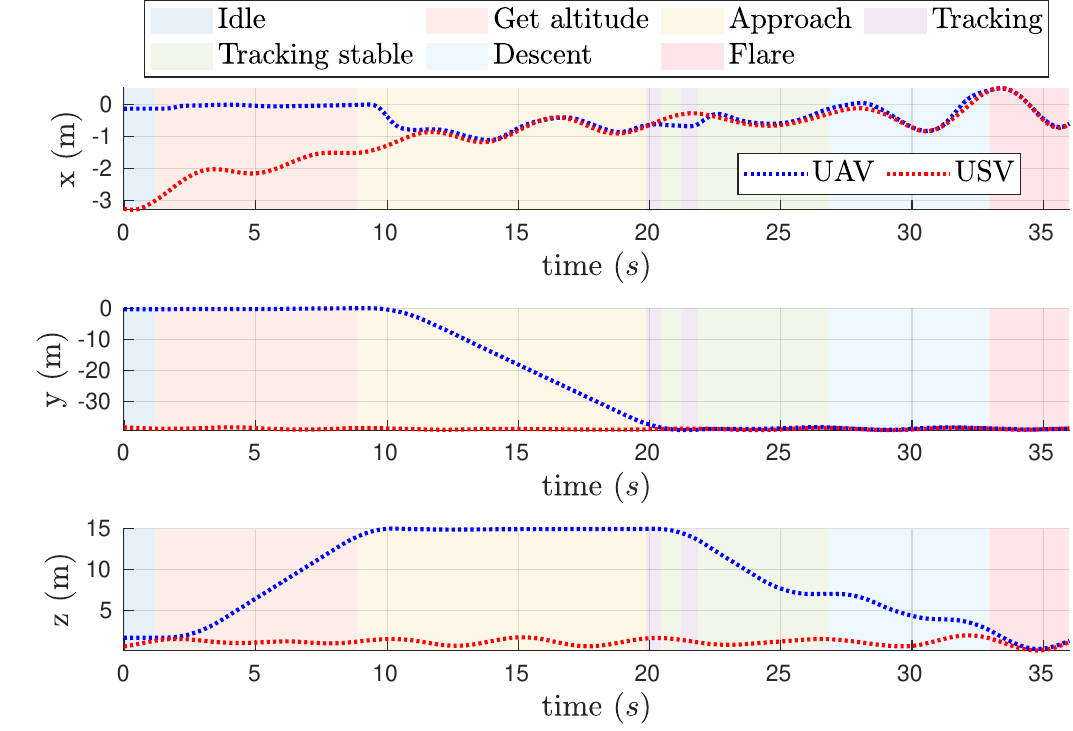}};
            \draw [amber] (8.55,0.63) -- (8.55,1.65);
            \draw [amber] (8.55,2.42) -- (8.55,3.42);
            \draw [amber] (8.55,4.23) -- (8.55,5.21);
        \end{tikzpicture}}
        \caption{Comparison of the UAV and USV positions together with the state machine states during one of the experiments on \textit{Moderate sea}. 
        The vertical yellow line indicates a touchdown.}
        \label{fig:position_comparison_waves}
    \end{figure}
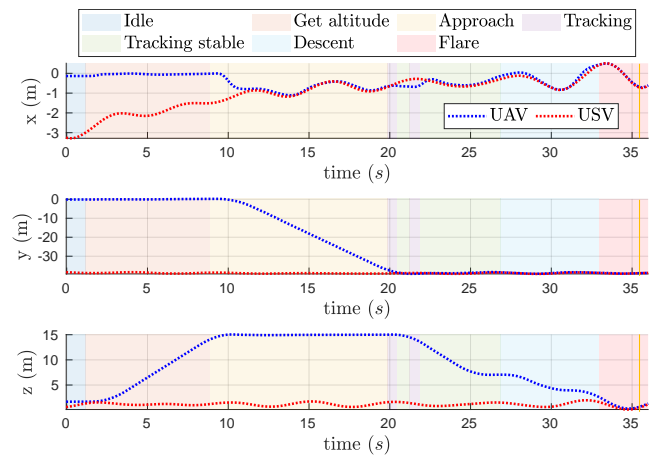
    Our approach was tested repeatably in realistic real-world environment simulations. 
    The takeoff position of the \ac{UAV} was placed randomly, always at least \SI{40}{\meter} away from the \ac{USV}.
    The trajectories of both the \ac{UAV} and \ac{USV} from one of the experimental trials are depicted in \autoref{fig:position_comparison_waves}.
    The graphs show that the \ac{UAV} ascended from the initial altitude \texttt{Idle} to the approach altitude. 
    It then began approaching the \ac{USV}'s position at time \SI{11}{\second} and continued through all flight states, which are detailed in \autoref{sec:mission_and_navigation}.
    The landing maneuver started at time \SI{29}{\second} and lasted approximately \SI{6}{\second} until the \texttt{Flare} state began.
    The touchdown was performed at time \SI{35.5}{\second}. 
    This shows that the proposed method is capable of generating trajectories that enable a fast approach and landing.
    Moreover, we show that our approach is able to land quickly within 6 seconds in waves with a height up to \SI{2.5}{\meter} and on an inclined \ac{USV}'s deck.

    To demonstrate the reliability of our method, we ran 100 randomly initiated landing tests.
    The accuracy of the landing is of the utmost importance, as it must be accurate in order to land successfully within the landing zone of the size of a roughly \SI{2}{\meter} x \SI{2}{\meter} deck.
    A significant roll or pitch deviation between the \ac{USV} and \ac{UAV} can result in overturning the \ac{UAV}, leading to potential damage and, thus, unsuccessful landing.
    On the other hand, maintaining the impact velocity within certain limits in order to prevent a strong impact with the \ac{USV}'s deck is crucial.

    \begin{figure*}[t!]
      \vspace{-0.3em}
      \centering
      \input{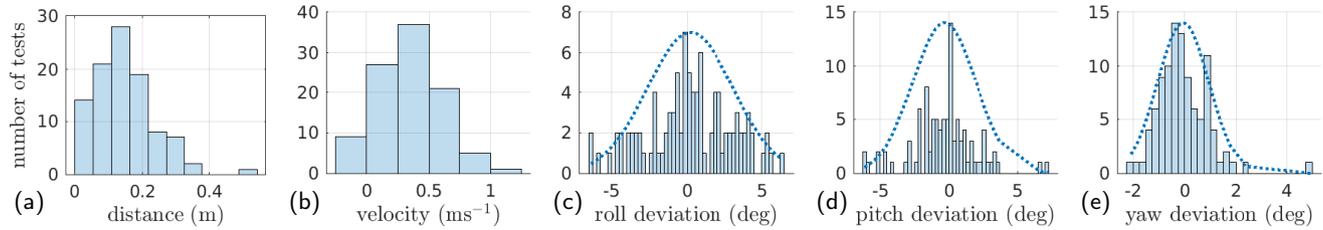}
      \caption{The assessment of the 100 \ac{UAV}'s touchdowns on \ac{USV}'s deck located in \textit{Moderate sea} conditions using normal distribution analysis for evaluation of the (a) position deviation $\mathcal{N}(0.15, 0.09^2)$, (b) horizontal velocity deviation $\mathcal{N}(0.36, 0.25^2)$, (c) roll deviation $\mathcal{N}(0.23, 2.85^2)$, (d) pitch deviation $\mathcal{N}(-0.32, 2.30^2)$ and (e) yaw deviation $\mathcal{N}(-0.05, 0.99^2)$  during touchdown.}
      \label{fig:landing_verification} 
      \vspace{-0.5cm}
    \end{figure*}  
    \begin{figure*}[t!]
      \vspace{-0.3em}
      \centering
      \input{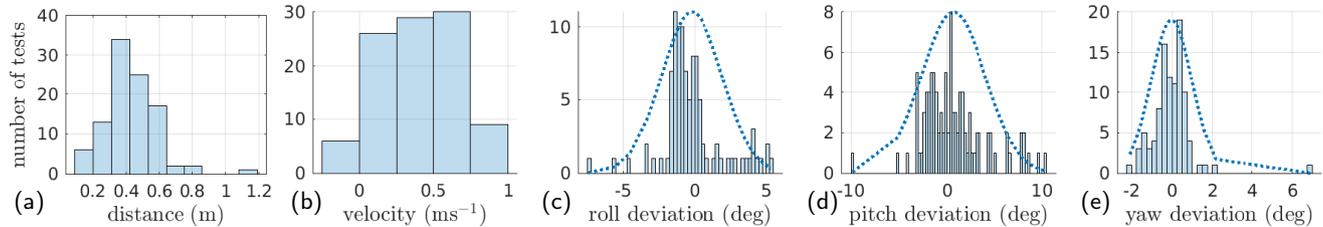}
      \caption{The results from 100 UAV touchdowns performed when the USV was following a path on \textit{Slight/Moderate sea}. (a) Position deviation $\mathcal{N}(0.44, 0.16^2)$, (b) horizontal velocity deviation $\mathcal{N}(0.41, 0.26^2)$, (c) roll deviation $\mathcal{N}(-0.14, 2.21^2)$, (d) pitch deviation $\mathcal{N}(0.59, 3.32^2)$ and yaw deviation $\mathcal{N}(-0.02, 1.05^2)$ during touchdown.}
      \label{fig:landing_moving_verification} 
      \vspace{-0.2cm}
    \end{figure*}  
    Statistical results evaluating the touchdown's precision, with normal distribution analysis, are depicted in \autoref{fig:landing_verification}.
    The errors in position during the touchdown are shown in \autoref{fig:landing_verification}a. 
    The maximum displacement recorded in one of the one hundred experiments was \SI{0.5}{\meter}, which is still considered as safe landing.
    All other experiments ended with a displacement smaller than \SI{0.35}{\meter}.
    The impact speed of the \ac{UAV} into the \ac{USV}'s deck (see 
    \autoref{fig:landing_verification}b)
    is within limits for safe landing. 
    The calculated velocity smaller than zero is caused by the inaccurate detection of the landing when the approach speed is close to zero.
    In \SI{91}{\percent} of the landing experiments, the roll angle deviation was within $\pm\SI{5}{\degree}$, and the pitch angle deviation did not exceed \SI{4}{\degree} in \SI{90}{\percent} of cases as shown in 
    \autoref{fig:landing_verification}c
    and in 
    \autoref{fig:landing_verification}d.
    Yaw angle control demonstrated high accuracy, with deviations exceeding two degrees only three times during all touchdowns (see 
    \autoref{fig:landing_verification}e).
    Therefore, it is worth mentioning that the yaw angle outlier at \SI{4.75}{\degree} was most likely caused by the \ac{USV}, which slightly rotated the \ac{UAV} before the touchdown was detected.
    
\subsection{USV following predefined path}\label{sec:usv_fol_pre_path}
    The proposed method was designed to be able to plan landing trajectories such that the \ac{USV} is not forced to stop or suspend its current task.
    Due to the increased complexity of landing on a moving boat, the simulated environment conditions were mitigated to \textit{Slight sea} and \textit{Moderate sea} levels.
    Consequently, the average wave height was around \SI{1.25}{\meter}.

    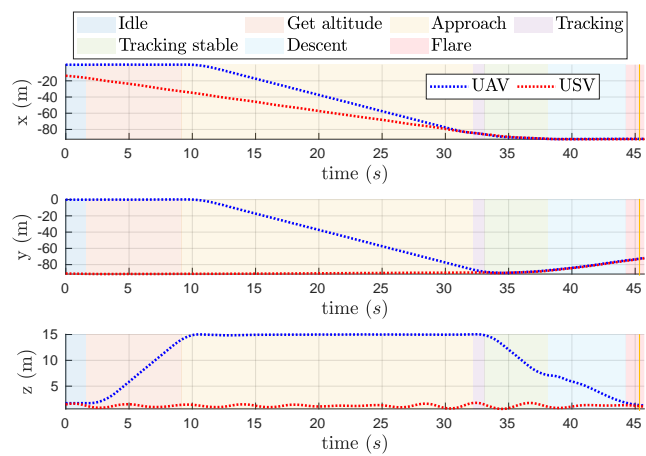
\begin{figure}[b!]
        \centering
        \subfloat{\begin{tikzpicture}[]
            \hspace{-0.3cm}
            \node[anchor=south west,inner sep=0] (a) at (0,0) { \includegraphics[clip, trim={30 226 64 215}, width = 0.5\textwidth]{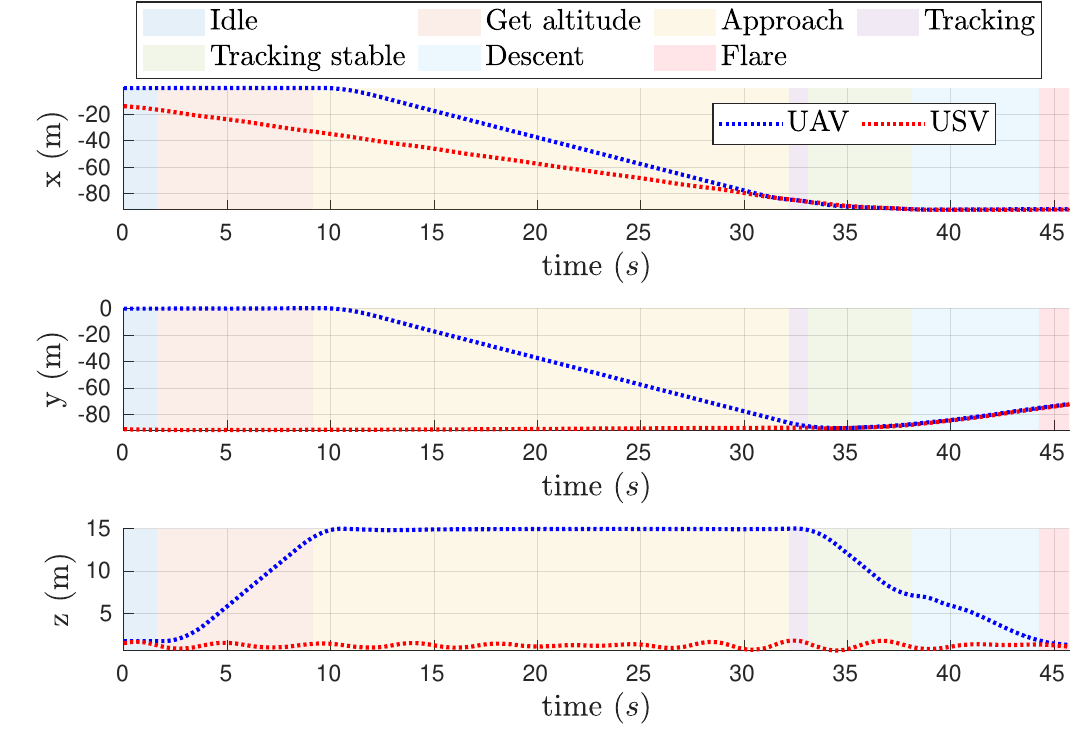}};
            \draw [amber] (8.55,0.63) -- (8.55,1.64);
            \draw [amber] (8.55,2.41) -- (8.55,3.42);
            \draw [amber] (8.55,4.18) -- (8.55,5.19);
        \end{tikzpicture}}
        \caption{Comparison of the UAV and USV positions together with the state machine states, while the boat was following a predefined path at a speed of \SI{2}{\meter\per\second}. 
        The vertical yellow line indicates a touchdown.}
        \label{fig:moving_boat_waves_position_comparison}
        \vspace{-0.2cm}
    \end{figure}
    One of the experiments in which the \ac{USV} followed a square-shaped path at a speed of around \SI{2}{\meter\per\second} is shown in \autoref{fig:moving_boat_waves_position_comparison}.
    The indicated velocity of the \ac{USV} is for information only because it was measured under \textit{Calm sea} conditions.
    However, in the tested sea conditions, the speed fluctuated due to the influence of waves.
    The experiment started with the \ac{UAV} approximately \SI{90}{\meter} away from the \ac{USV}.
    The \ac{UAV} was able to catch up with the \ac{USV} at time \SI{32}{\second} and began descending into the \texttt{Follow} phase.
    At \SI{38}{\second}, as can be seen from the attached \autoref{fig:moving_boat_waves_position_comparison}, the \ac{USV} changed its direction of motion.
    Nevertheless, all conditions for landing were met and the landing maneuver started almost immediately, continuing into the \texttt{Flare} state.
    The \ac{UAV} landed at \SI{45}{\second}.

    The aforementioned experiment was conducted 100 times with \SI{100}{\percent} success rate to gather statistical results (see \autoref{fig:landing_moving_verification}), showing slightly worse results compared to landing on a \ac{USV} being carried by the current (\autoref{sec:carried_by_current}).
    The measured deviations upon touchdown are higher here due to the increased accelerations experienced by the \ac{USV}, resulting in a less accurate estimation of the \ac{USV}'s motions.
    The mean position error of touchdown increased from \SI{15}{\centi\meter} to \SI{0.44}{\meter} (see 
    \autoref{fig:landing_moving_verification}a).
    On the other hand, the mean approach velocity increased only by \SI{0.05}{\meter\per\second}, as can be seen in 
    \autoref{fig:landing_moving_verification}b.
    The yaw's deviation mean value (
    \autoref{fig:landing_moving_verification}e)
    is close to zero, and most tests ended with a yaw deviation between $\pm 2$ degrees.
    However, values larger than \SI{2}{\degree} also appeared, which was caused by the fact that some landing attempts were performed during the \ac{USV}'s turning maneuver, where the \ac{USV}'s yaw angle estimation was lagging slightly behind.
    The ability to align the roll and pitch angles upon the touchdown ensured that most of the touchdown attempts ended up close to the demanded attitude (see 
    \autoref{fig:landing_moving_verification}c and
    \autoref{fig:landing_moving_verification}d,
    ensuring a safe touchdown.

\subsection{Comparison with state of the art}\label{state_of_the_art_comparison}
    According to our knowledge and based on the comprehensive state-of-the-art study (for the summary, see \autoref{tab:comparison_stoa}), we found as the most relevant work of \cite{gupta2022landing}.
    However, the method MPC-NE in \cite{gupta2022landing} utilizes only the AprilTag measurements for the \ac{USV}'s position estimation and does not align the yaw angle of the \ac{UAV} with the \ac{USV}'s. 
    MPC-NE is not capable of the initial approach to the \ac{USV}'s position when the \ac{USV} is not in the \ac{FOV} of the \ac{UAV}'s camera.  
    Therefore, the landing phase is the only part of our method MPC-FAA that can be compared with the state-of-the-art method MPC-NE.
    \begin{figure*}[t!]
        \begin{subcolumns}[0.3\textwidth]
        \centering
        \subfloat[Illustration of the touchdown's positions.]{
            \includegraphics[clip, trim={40 0 60 30},width = \subcolumnwidth]{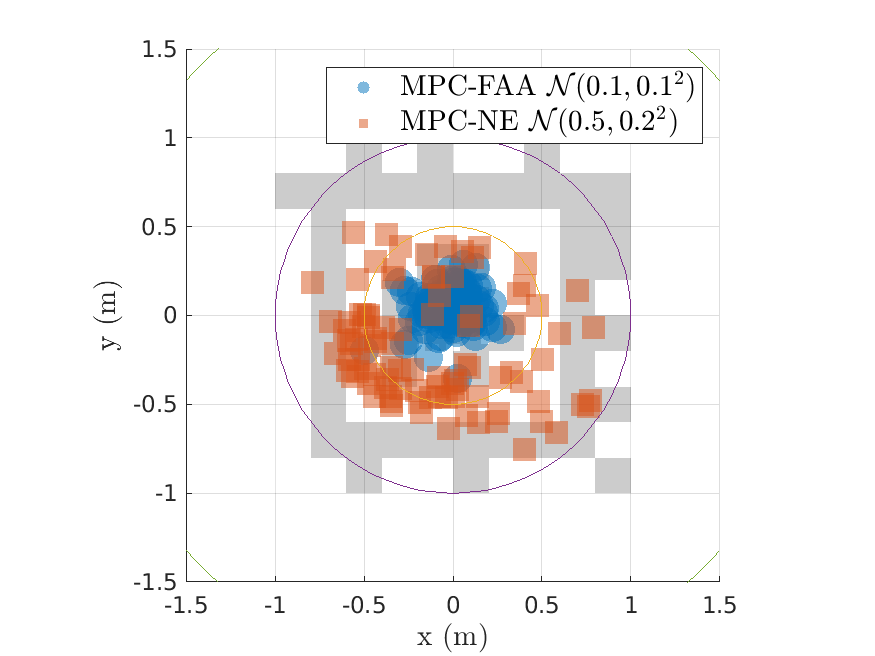}
            \label{fig:parakh_precision_comparison_05}
        }
        \nextsubcolumn  
        \subfloat[Vertical touchdown velocities during the landing.]{
            \includegraphics[clip, trim={10 0 30 10},width = \subcolumnwidth]{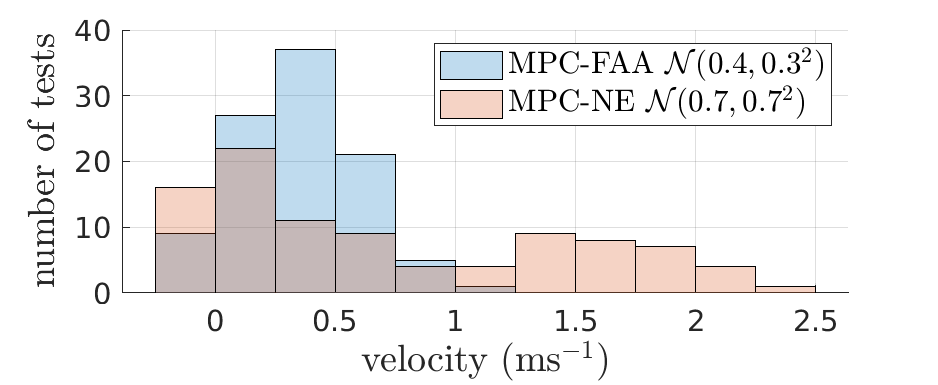}
            \label{fig:parakh_approach_velocity_comparison}
        }\nextsubfigure
        \subfloat[Landing maneuver duration.]{
            \includegraphics[clip, trim={10 0 30 10},width = \subcolumnwidth]{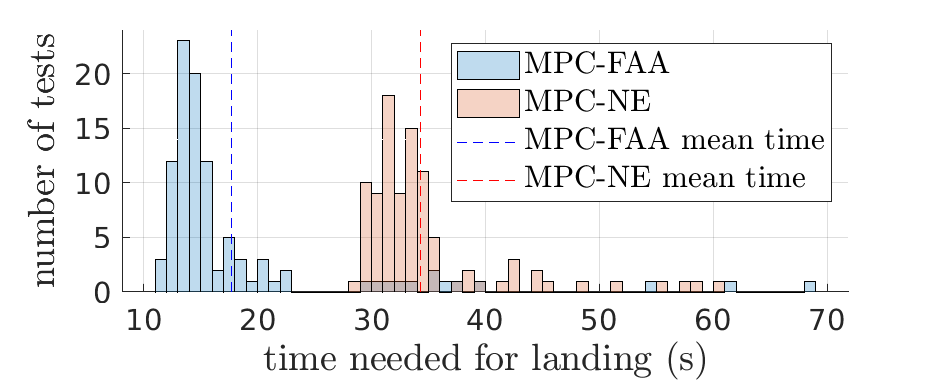}
            \label{fig:parakh_time_comparison}
        }\nextsubcolumn 
        \subfloat[Roll deviation during the touchdown.]{
            \includegraphics[clip, trim={10 0 30 10},width = \subcolumnwidth]{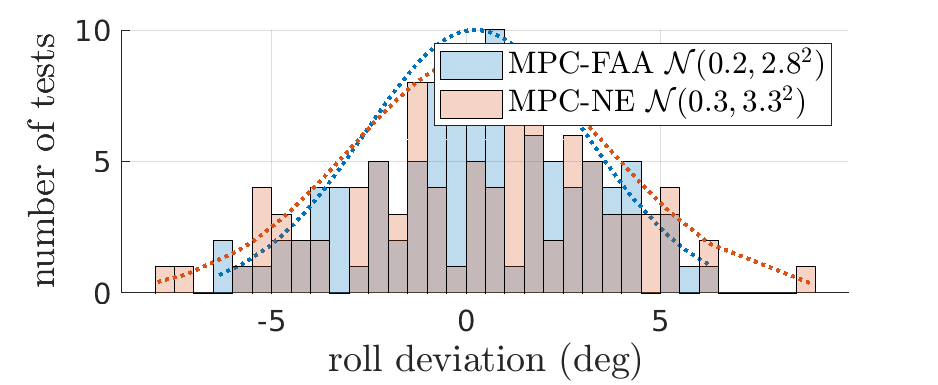}
            \label{fig:parakh_roll_comparison}
        }\nextsubfigure
        \subfloat[Pitch deviation during the touchdown.]{
            \includegraphics[clip, trim={10 0 30 9},width = \subcolumnwidth]{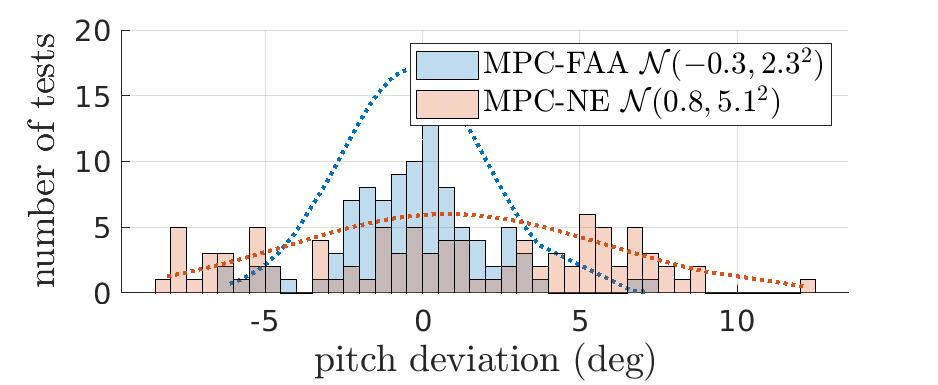}
            \label{fig:parakh_pitch_comparison}
        }
        \end{subcolumns}
        \caption{Comparison of the proposed MPC-FAA and MPC-NE methods in waves with a height of up to 2.5 m (\textit{Moderate sea}).
        }
        \label{fig:parakh_comparison_05}
        \vspace{-0.5cm}
    \end{figure*}
    
    More than one thousand simulations for each method were done to obtain the statistical results.
    The primary focus of interest while comparing both methods was the accuracy of the landing, specifically, the precision of position, attitude, and impact velocity.
    Another crucial parameter that was compared is the duration of the landing maneuver itself.

\subsubsection{Comparison of landing precision on \textit{Moderate sea}}\label{sec:landing_precision_comparison_on_moderate_sea}
    We first compare MPC-FAA with MPC-NE in scenarios close to \textit{Moderate sea} conditions.
    Touchdown positions accuracy of both methods shown in \autoref{fig:parakh_precision_comparison_05} is fitted to the normal distribution, which expresses the mean value of the distance from the center of the landing platform (AprilTag) with the standard deviation.
    The touchdown success rate of MPC-NE method was \SI{95}{\percent}, whereas our method MPC-FAA showed \SI{100}{\percent} success rate.
    Moreover, MPC-FAA method achieved a position deviation 3.5 times smaller than MPC-NE method.
    The comparison of the vertical approach velocity during the touchdown, shown in \autoref{fig:parakh_approach_velocity_comparison}, displays that the method MPC-FAA, unlike MPC-NE, ensured that the impact velocity was not higher than \SI{1}{\metre\per\second}.
    
    The graph in \autoref{fig:parakh_roll_comparison} shows that MPC-FAA and MPC-NE methods were able to demonstrate accuracy in synchronizing the \ac{UAV}'s roll angle with the \ac{USV}'s.
    Nevertheless, MPC-FAA achieved a better standard deviation \SI{2.85}{\degree} compared to MPC-NE with \SI{3.35}{\degree}.
    MPC-FAA outperformed MPC-NE with a 2 times smaller standard deviation for the pitch angle deviation (see \autoref{fig:parakh_pitch_comparison}).

    Out of the 95 successful landings achieved by MPC-NE, the landing maneuver was performed within \SI{60}{\second} in all cases (see \autoref{fig:parakh_time_comparison}).
    However, the average landing time was \SI{34}{\second}, which is 2 times longer than what was achieved by MPC-FAA method.
    At the same time, as shown in \autoref{fig:parakh_precision_comparison_05}, our approach consistently achieves a higher level of accuracy when landing in \textit{Moderate sea} conditions.

\subsubsection{Comparison of landing precision in varied sea conditions}\label{sec:landing_comparison_on_varied_sea_conditions}
    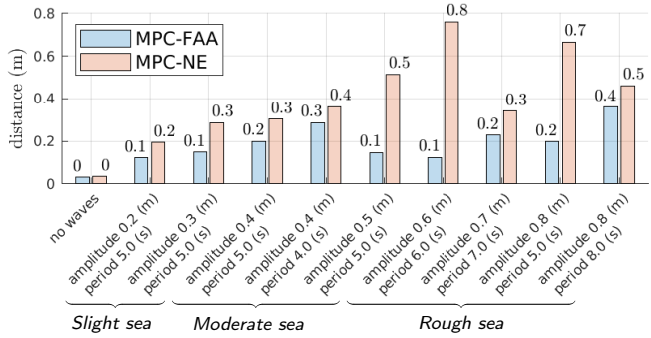
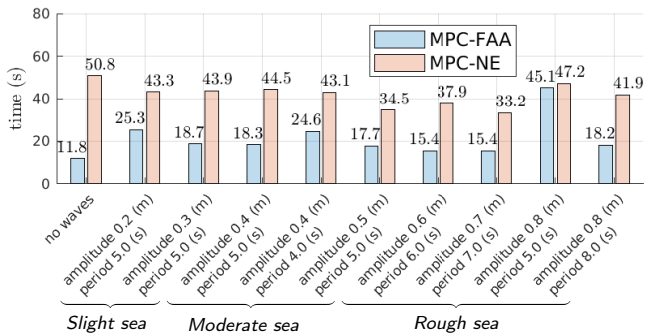
\begin{figure}[t!]
        \centering
        \vspace{-0.3cm}
        \subfloat[Landing precision as a distance from landing pad center.]{\begin{tikzpicture}[]
        \node[anchor=south west,inner sep=0] (a) at (0,0) {\includegraphics[clip, trim={25 0 40 0}, width=0.49\textwidth]{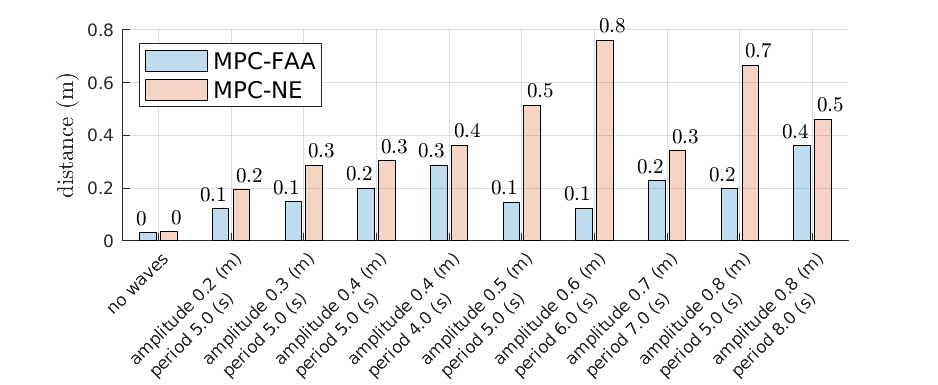}
        \label{fig:mean_position_diff}};
        \draw [decorate, decoration = {calligraphic brace}] (7.5,0) --  (4.5,0);
        \draw (6,-0.3) node [text=black, opacity=1] {\scriptsize \textit{Rough sea}};
        \draw [decorate, decoration = {calligraphic brace}] (4.4,0) --  (2.2,0);
        \draw (3.2,-0.3) node [text=black, opacity=1] {\scriptsize \textit{Moderate sea}};
        \draw [decorate, decoration = {calligraphic brace}] (2.1,0) --  (0.8,0);
        \draw (1.4,-0.3) node [text=black, opacity=1] {\scriptsize \textit{Slight sea}};
        \end{tikzpicture}}
        \hfill
        \vspace{-0.25cm}
        \subfloat[Landing maneuver duration.]{\begin{tikzpicture}[]
        \node[anchor=south west,inner sep=0] (a) at (0,0) {
        \includegraphics[clip, trim={25 0 40 0},width=0.49\textwidth]{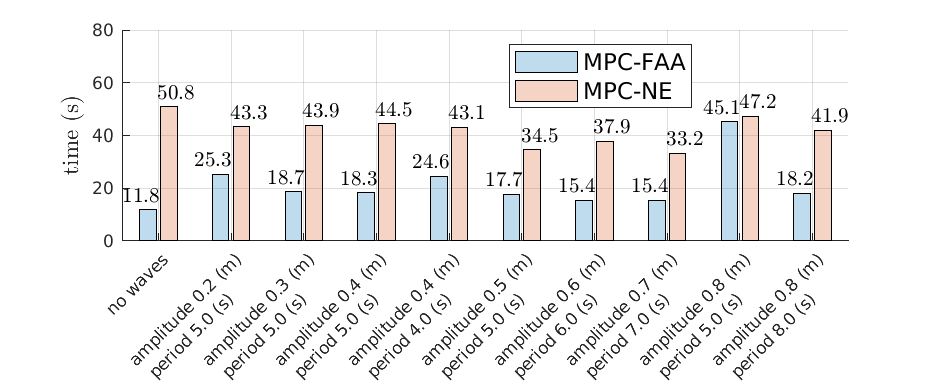}\label{fig:mean_landing_maneuver_time}};
        \draw [decorate, decoration = {calligraphic brace}] (7.5,0) --  (4.5,0);
        \draw (6,-0.3) node [text=black, opacity=1] {\scriptsize \textit{Rough sea}};
        \draw [decorate, decoration = {calligraphic brace}] (4.4,0) --  (2.2,0);
        \draw (3.2,-0.3) node [text=black, opacity=1] {\scriptsize \textit{Moderate sea}};
        \draw [decorate, decoration = {calligraphic brace}] (2.1,0) --  (0.8,0);
        \draw (1.4,-0.3) node [text=black, opacity=1] {\scriptsize \textit{Slight sea}};
        \end{tikzpicture}}
        \caption{Comparison of landing maneuver precision and landing duration of the proposed MPC-FAA method and state-of-the-art MPC-NE approach across individual scenario setups.}
        \label{fig:position_diff_time}
        \vspace{-0.3cm}
    \end{figure}
    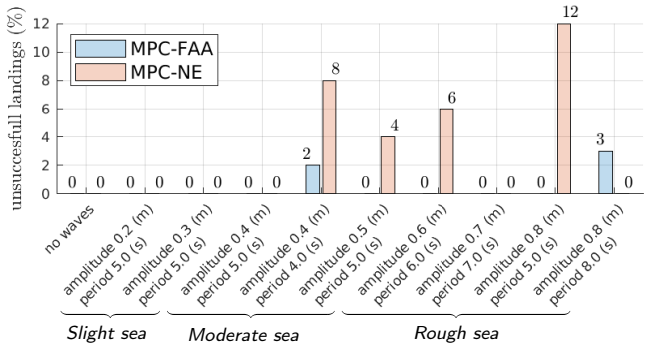
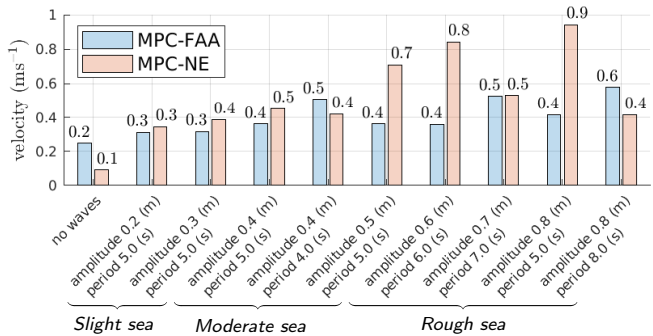
\begin{figure}[t!]
        \centering
        \vspace{-0.1cm}
        \subfloat[Number of unsuccessful landings from 100 runs.]{\begin{tikzpicture}[]
        \node[anchor=south west,inner sep=0] (a) at (0,0) {\includegraphics[clip, trim={25 0 40 0}, width=0.49\textwidth]{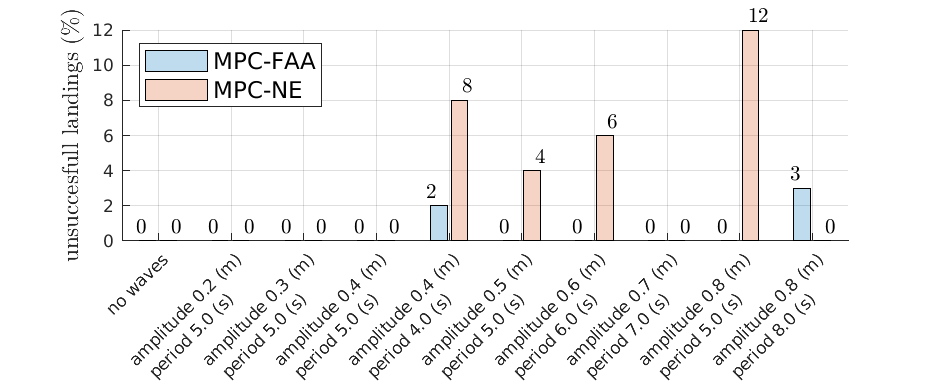}
        \label{fig:unsuccess_rate}};
        \draw [decorate, decoration = {calligraphic brace}] (7.5,0) --  (4.5,0);
        \draw (6,-0.3) node [text=black, opacity=1] {\scriptsize \textit{Rough sea}};
        \draw [decorate, decoration = {calligraphic brace}] (4.4,0) --  (2.2,0);
        \draw (3.2,-0.3) node [text=black, opacity=1] {\scriptsize \textit{Moderate sea}};
        \draw [decorate, decoration = {calligraphic brace}] (2.1,0) --  (0.8,0);
        \draw (1.4,-0.3) node [text=black, opacity=1] {\scriptsize \textit{Slight sea}};
        \end{tikzpicture}}
        \hfill
        \vspace{-0.25cm}
        \subfloat[Touchdown velocity during the landing.]{\begin{tikzpicture}[]
        \node[anchor=south west,inner sep=0] (a) at (0,0) {\includegraphics[clip, trim={25 0 40 0},width=0.49\textwidth]{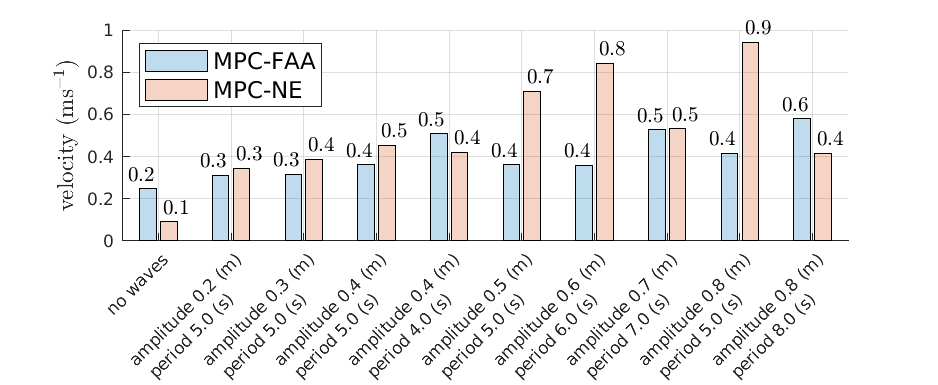}\label{fig:mean_approach_velocity}};
        \draw [decorate, decoration = {calligraphic brace}] (7.5,0) --  (4.5,0);
        \draw (6,-0.3) node [text=black, opacity=1] {\scriptsize \textit{Rough sea}};
        \draw [decorate, decoration = {calligraphic brace}] (4.4,0) --  (2.2,0);
        \draw (3.2,-0.3) node [text=black, opacity=1] {\scriptsize \textit{Moderate sea}};
        \draw [decorate, decoration = {calligraphic brace}] (2.1,0) --  (0.8,0);
        \draw (1.4,-0.3) node [text=black, opacity=1] {\scriptsize \textit{Slight sea}};
        \end{tikzpicture}}
        \caption{Comparison of the number of unsuccessful landings and touchdown velocity during the landing across individual scenario setups of the proposed method MPC-FAA and state-of-the-art MPC-NE approach.}
        \label{fig:unsuccess_approach_velocity}
        \vspace{-0.4cm}
    \end{figure}
    To comprehensively compare MPC-FAA method and MPC-NE method in different conditions, we selected 10 different environment setups that cover a wide range of possible offshore conditions.
    The simulated scenarios differ in wave amplitude and period, such that the simulated sea varied from \textit{Calm sea} to \textit{Rough sea} (see \autoref{tab:wave_definition}).
    Both MPC-FAA and MPC-NE methods were tested for every scenario one hundred times. 
    The results are shown in (\autoref{fig:position_diff_time} -- \autoref{fig:unsuccess_approach_velocity}), where the x-axis labels represent the wave amplitude and period used in simulated scenarios, as described in \autoref{sec:sim_environment}.

    The comparison of the touchdown position precision is shown in \autoref{fig:mean_position_diff}.
    According to this graph, MPC-FAA achieved a mean touchdown position closer to the center of the \ac{USV}'s deck across all tested scenarios when compared to MPC-NE.
    Our method's largest average position deviation (\SI{37}{\centi\meter}) occurred in a challenging scenario with amplitude \SI{0.8}{\meter} and a period of \SI{8}{\second}, where the waves reached \SI{5}{\meter} in height.
    On the other hand, MPC-NE has the worst touchdown precision when the wave's amplitude was set to \SI{0.6}{\meter} and the period to \SI{6}{\second}.
    The worst position deviation of MPC-NE is most probably due to the fact that the method neglects the sway and surge movements while the waves affect the \ac{USV}'s motion in both vertical and horizontal directions.

    \autoref{fig:mean_landing_maneuver_time} compares the average duration of the landing maneuvers.
    Thanks to MPC-FAA method's ability to use estimation of the \ac{USV}'s movements before the \ac{UAV} reached its position, it was ensured that the landing was, on average, 2 times faster than MPC-NE method.
    The method MPC-NE needs longer observation of the \ac{USV}'s movements for the purpose of identifying the waves.
    Moreover, due to neglecting surge and sway movements, \ac{UAV} may lose sight of the AprilTag, which leads to a termination of the landing maneuver. 
    Based on the data presented in \autoref{fig:mean_landing_maneuver_time}, most landing interruptions of our method occurred in a scenario where the amplitude of the waves was \SI{0.8}{\meter} and the period was \SI{5}{\second}.
    In such a scenario, our method had an average landing duration of around \SI{45}{\second}. 
    However, there was no significant increase in the landing time for the method MPC-NE.
    
    Both methods achieved a low unsuccess rate of touchdowns, as can be seen in \autoref{fig:unsuccess_rate}.
    MPC-FAA method failed only 5 times, while MPC-NE method failed 30 times.
    The comparison of the vertical approach velocity (see \autoref{fig:mean_approach_velocity}) shows that MPC-FAA method is consistent over all tested scenarios.
    On the other hand, MPC-NE method ensured a softer landing in scenarios without waves but a more aggressive and dangerous landing in high-wave scenarios.
    Overall, the proposed method MPC-FAA outperformed MPC-NE in almost every aspect compared and showed that the MPC-FAA is reliable in various sea conditions up to \textit{Rough sea} (wave height 4 m).

\section{REAL-WORLD EXPERIMENTS}
    The real-world experiments were conducted in freshwater reservoirs, so to induce conditions similar to those tested in the simulation, it was necessary to create artificial waves.
    This ensures the repeatability of the experiments over a long time horizon, regardless of the weather, which is responsible for creating the waves on the open waters.
    The lightweight inflatable mockup of the \ac{USV} with a landing deck was used for easy manipulation, allowing for the creation of artificial waves with different amplitudes and frequencies.
    \begin{figure*}[t!]
        \centering
        \vspace{-0.4cm}
        \input{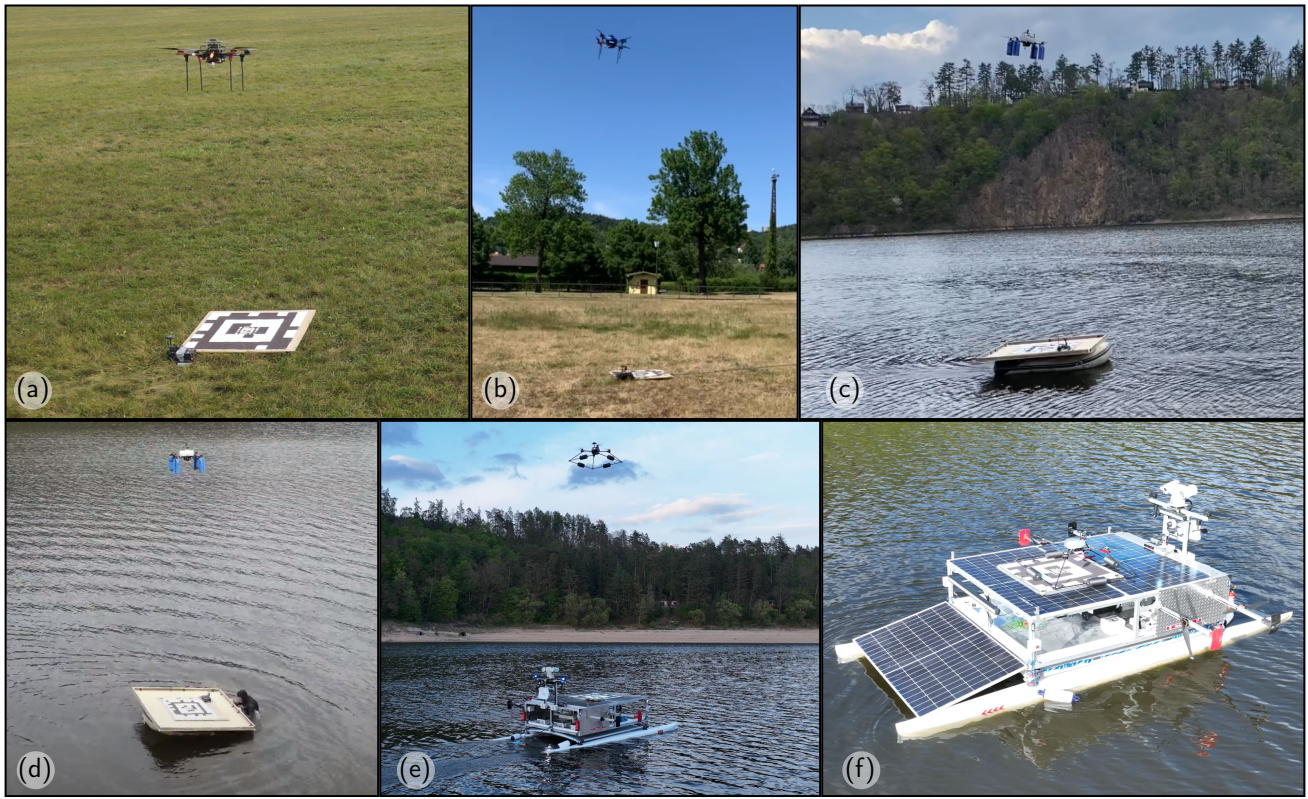}
        \caption{Captured photographs from real-world experiments showcasing the method from an initial approach until a precise touchdown. The photos show 
        (a) landing the UAV on the landing pad on land, 
        (b) tracking a moving landing pad with the UAV on land,
        (c) tracking the USV with the UAV in high wind,
        (d) tracking the USV with the UAV in high wind and artificial waves,
        (e) tracking moving USV with the UAV, and
        (f) UAV successfully landed on the USV. \url{https://youtu.be/7Cvdttz9ZZk}}
        \label{fig:real_world_pictures}
        \vspace{-0.4cm}
    \end{figure*}

    \subsection{UAV and USV platforms}
    The \ac{UAV} was built using the Tarot T650 frame with \acl{BLDC} Tarot 4114 320KV motors and Turnigy Multistar 51A \aclp{ESC}. 
    The \ac{UAV} was equipped with a Pixhawk 4 autopilot with \ac{GPS}, and powered by a LiPo 6S \SI{8000}{{\milli\ampere\hour}} battery. 
    The proposed method runs onboard, together with the rest of the autonomy stack, on NUC8i7BEH high-level computer with an i7 8559U processor, 16GB of DDR4 RAM, and a 500GB SSD.
    The details of the \ac{UAV} platform are in \cite{MRS2022ICUAS_HW}, however, some modifications were needed due to the \ac{UAV}'s above-water operation.
    The \ac{UAV} was equipped with the RealSense D435 camera for detection of the AprilTag.
    Furthermore, the mvBlueFOX MLC200wG camera was used to observe the \ac{UV} \acp{LED} that are placed on the \ac{USV}'s deck.
    The \ac{UAV} was equipped with floats on its legs, which ensures the \ac{UAV} can float in case it lands on the water surface.
    All electronic components were also protected with a water-resistant spray and placed in the 3D-printed water-proofed cover.

    For conducting the first experiments, the landing pad was mounted on an inflatable raft towed behind a manned vessel.
    This mockup, towed by the manned vessel, was later exchanged for the unmanned \ac{USV} (see \autoref{fig:real_world_pictures}f).
    The landing platform was designed in the shape of a square with a side length of \SI{2}{\meter}.
    The Boat unit, which contains a NUC8i7BEH computer, LiPo battery pack, \ac{GPS}, \ac{IMU} units, and other electronics, was mounted on the top of a deck.
    The landing area with a side length of \SI{80}{\centi\meter} was covered with AprilTags and equipped with \ac{UV} \acp{LED} blinking marker used for relative localization of the \ac{USV} from the \ac{UAV} onboard using AprilTag and \ac{UVDAR} detectors.

    It is also worth mentioning that even after a successful touchdown, the \ac{UAV} can be flipped or dropped from the \ac{USV}’s deck due to the continuous movements of the \ac{USV} in the waves.
    To ensure that the \ac{UAV} remains stable on the deck, the magnetic adhesion device presented in \cite{zhang2024precise} can be used to hold the \ac{UAV} after touchdown.
    Alternatively, a suitable solution based on docking station~\cite{drones6010017} can be used.

\subsection{Real-world landing on USV}
\begin{figure*}[t!]
        \centering
        \input{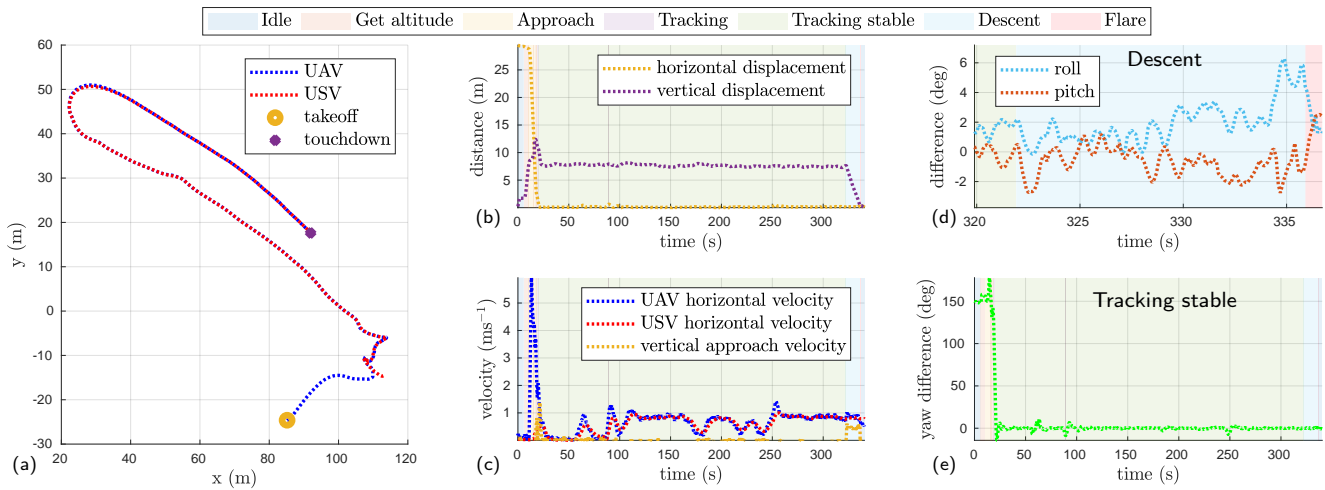}
        \caption{Evaluation of the USV following and landing on its deck. Individual graphs depict (a) UAV's and USV's path with marked takeoff and touchdown positions, (b) position displacements between UAV and USV, (c) horizontal velocity together with vertical approach velocity, (d) the difference between the roll and pitch of the UAV and USV during \texttt{Descent} and \texttt{Flare} states, and (e) yaw difference between UAV and USV.}
        \label{fig:real_world_following}
        \vspace{-0.3cm}
    \end{figure*}

    The results from one of the complex real-world experiments are shown in \autoref{fig:real_world_following}, where the comparison of the \ac{UAV}'s and \ac{USV}'s paths in a 2D plane is shown in \autoref{fig:real_world_following}a.
    The initial segment of the \ac{USV}'s trajectory begins in shallow water near the shore, where we artificially created erratic movements to verify the robustness and precision of the tracking part of MPC-FAA method.
    The \ac{USV} is followed by the \ac{UAV} at the tracking height, where the mutual position is shown in \autoref{fig:real_world_following}b and velocity in \autoref{fig:real_world_following}c.
    The velocity was not constant to emulate the \ac{USV}'s movements in a wave environment, allowing us to validate the capability of our method to track the \ac{USV} even under varying speeds.

    The transition to the LANDING phase was intentionally delayed to demonstrate the method's capability to track reliably \ac{USV} even while turning, thus the landing started at \SI{324}{\second} from takeoff.
    In the LANDING phase, especially during the touchdown itself, aligning the attitude angles is crucial.
    Therefore, \autoref{fig:real_world_following}d illustrate the differences between \ac{UAV}'s and \ac{USV}'s attitude angles focused on \texttt{Descent} and \texttt{Flare} states, where our system performs roll and pitch alignment.
    In contrast, the heading alignment is carried out from the FOLLOW phase, as shown in \autoref{fig:real_world_following}e.
    The touchdown was executed with a position deviation of approximately \SI{20}{\centi\meter} and an attitude deviation smaller than \SI{2}{\degree} in every axis.
    \begin{figure}[t!]
        \centering
        \vspace{0.2cm}
        \subfloat[A comparison of the x,y, and z coordinates of the UAV and USV.]{\begin{tikzpicture}[]
        \node[anchor=south west,inner sep=0] (a) at (0,0) {
        \includegraphics[clip, trim={45 230 48 215}, width=0.49\textwidth]{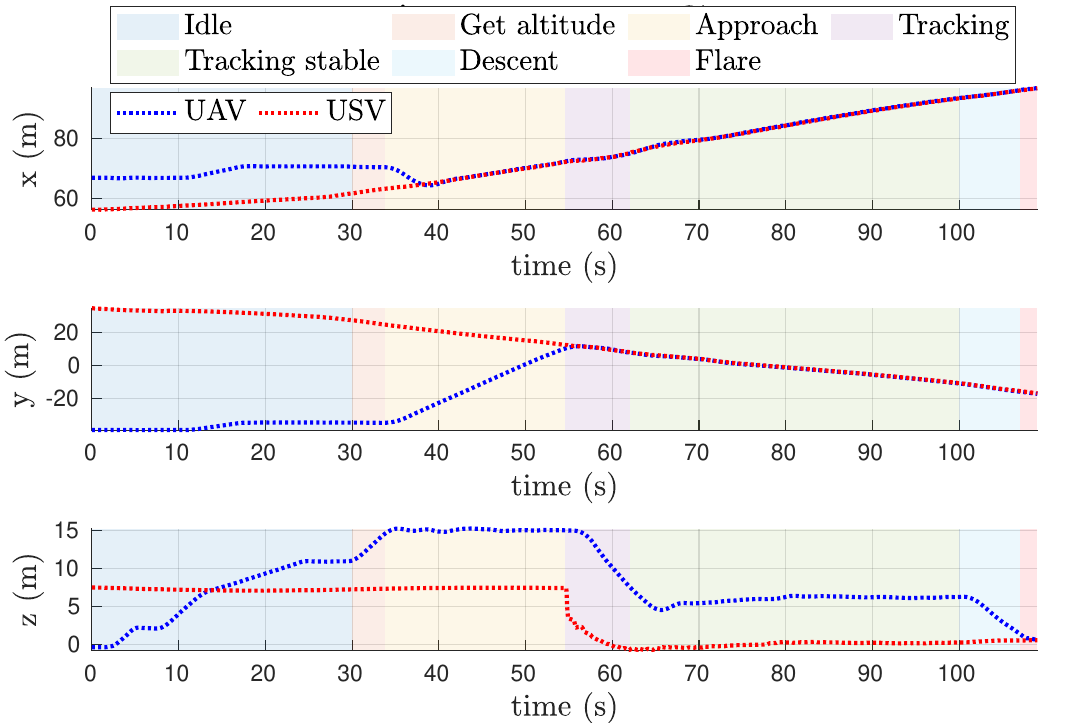}};
        \draw [amber] (8.13,0.6) -- (8.13,1.6);
        \draw [amber] (8.13,2.32) -- (8.13,3.28);
        \draw [amber] (8.13,4.04) -- (8.13,5.05);
        \end{tikzpicture}
        }
        \vspace{-0.2cm}
        \hfill
        \subfloat[USV's roll and pitch angles, and the difference between the UAV's and USV's angles during the landing.]{\begin{tikzpicture}[]
        \node[anchor=south west,inner sep=0] (a) at (0,0) {
        \includegraphics[clip, trim={45 230 48 220},width=0.49\textwidth]{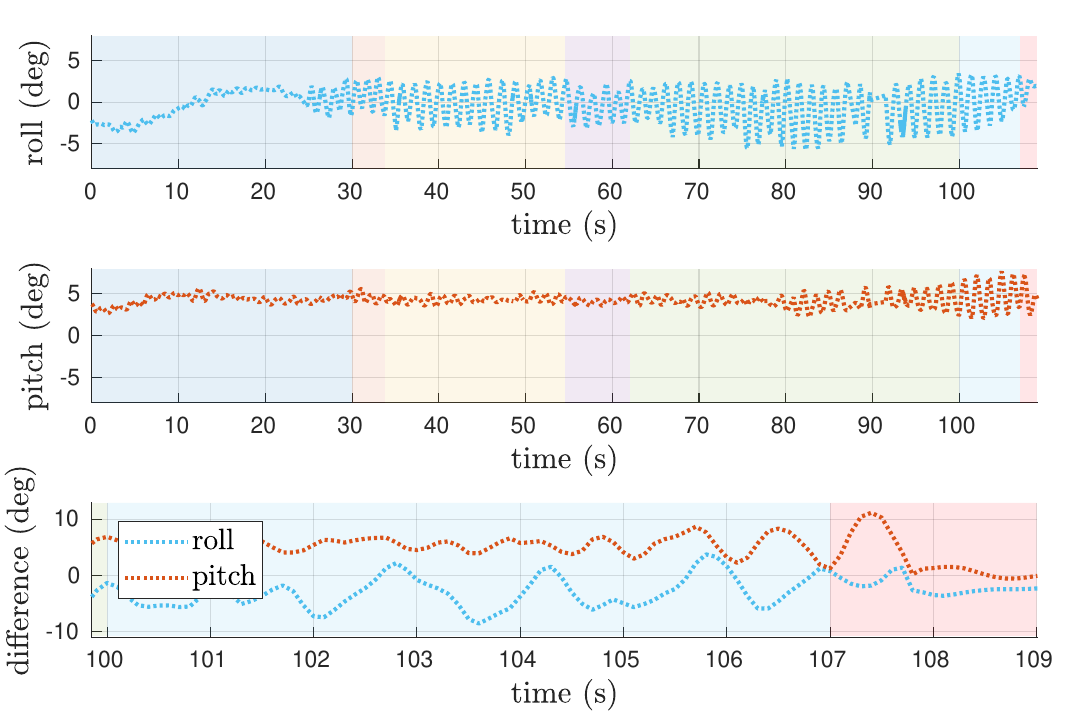}};
        \draw [amber] (7.25,0.59) -- (7.25,1.65);
        \draw [amber] (8.13,2.42) -- (8.13,3.5);
        \draw [amber] (8.13,4.27) -- (8.13,5.35);
        \end{tikzpicture}}
        \caption{Evaluation of the real-world landing experiment with artificial waves. 
        The vertical yellow line indicates a touchdown. \url{https://youtu.be/U29zBVzZ3sI}}
        \label{fig:uav_usv_position_comparison_real_world}
        \vspace{-0.5cm}
    \end{figure}
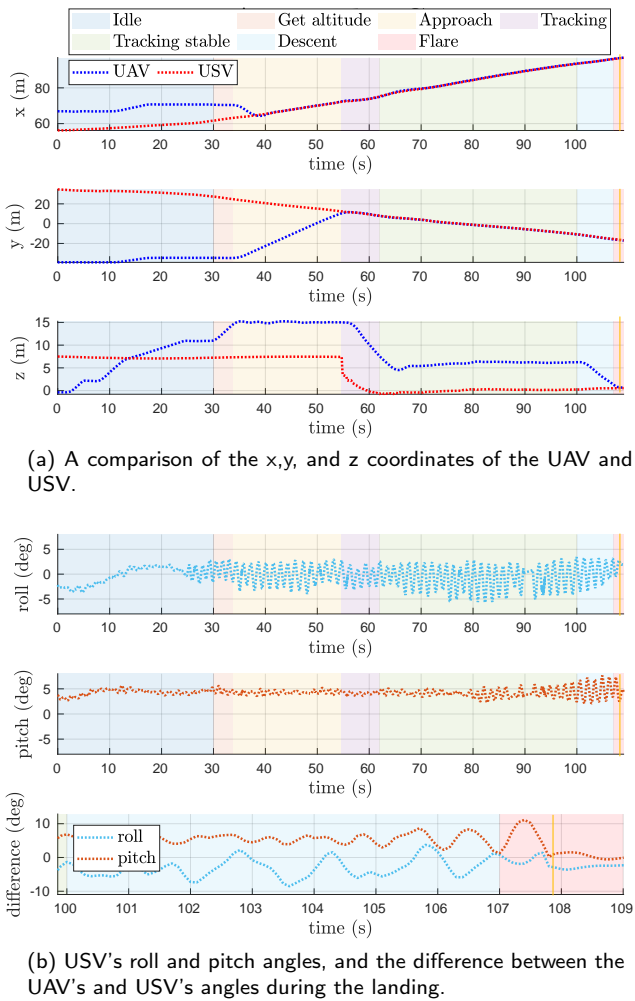
    
    The next experiment summarized in \autoref{fig:uav_usv_position_comparison_real_world} differs from the previous one by the presence of wind with speed of \SI{7}{\kilo\meter\per\hour}, wind gusts up to \SI{17}{\kilo\meter\per\hour} but mainly due to the artificial waves.
    The comparison of the \ac{UAV}'s and \ac{USV}'s positions over time is shown in \autoref{fig:uav_usv_position_comparison_real_world}a.
    In this experiment, the pilot controlled the first \SI{30}{\second} of the flight to simulate real-world deployment, where the drone is initially controlled by a different program, such as in the motivation scenario, where an exploration planner is used to search for garbage.
    Afterward, the landing algorithm is started to land from the current location.
    The \ac{UAV} was guided from its initial position to the above-water surface.
    In the first 5 seconds after initiating \ac{MPC}-based trajectory generator, the \ac{UAV} gained an altitude $h_a = \SI{15}{\meter}$ and began an approach to the \ac{USV}'s position. 
    When the \ac{UAV} obtained data from the AprilTag detection system, the estimation of the \ac{USV}'s z position was corrected.
    This phenomenon, which occurred between 55 and 60 seconds from the start, clearly indicates that relying solely on \ac{GPS} data for estimating the \ac{USV}'s altitude is insufficient and that the estimated altitude can not be used for a safe landing.
    The conditions for the final landing were met shortly after the tracking of the \ac{USV} started.
    
    Throughout the entire experiment, the \ac{USV} was artificially rocked, resulting in a maximum pitch angle of \SI{8}{\degree} and minimum roll angle of \SI{-6}{\degree} (see \autoref{fig:uav_usv_position_comparison_real_world}b).
    The LANDING phase lasted approximately \SI{8}{\second} and was concluded with a successful touchdown with attitude angles deviation lower than \SI{2}{\degree} and position deviation approximately \SI{15}{\centi\meter}.
    The effectiveness of \ac{FAA} in the reduction of the difference between roll and pitch angles during the flare maneuver is shown in \autoref{fig:uav_usv_position_comparison_real_world}c.
    Despite the significant deviation at the beginning of the Flare state, the increasing penalization of the \ac{UAV}'s attitude effectively reduces this deviation to a minimum until the time of touchdown.
    During these experiments, we verified that the proposed method can operate in a real-world environment with almost identical behavior to that of simulations.

\section{Conclusions}\label{sec:conclusion}
    In this paper, we proposed a novel method for autonomous trajectory generation of \ac{UAV} landing on top of a moving and rocking \ac{USV}.
    The novel \ac{MPC}-based trajectory generation relies on the accurate estimation and prediction of \ac{USV} states, along with precise autonomous control of a \ac{UAV}.
    Thanks to the dynamic modification of the penalization matrices, the vertical approach speed is reduced, and attitude angles are aligned with the \ac{USV} to decrease the impact of the \ac{UAV}'s gear just before the touchdown.
    The \ac{UAV} can successfully land in \textit{Moderate sea} conditions or even on \textit{Rough sea} conditions when using our approach.
    MPC-FAA achieved a reliability rate of \SI{99.5}{\percent} in harsh conditions in simulated cases, whereas the state-of-the-art approach had 6 times more unsuccessful touchdowns.
    Moreover, the results of the proposed system indicate a substantial improvement in touchdown accuracy, achieving approximately twice better precision while also completing the landing maneuver \SI{50}{\percent} faster than MPC-NE method. 
    These outcomes highlight the efficiency of our approach in comparison to existing techniques.
    Moreover, the MPC-based trajectory generator for landing the \ac{UAV} on the \ac{USV} was verified by conducting numerous real-world experiments.










\bibliographystyle{cas-model2-names}

\bibliography{cas-refs}





\begin{acronym}
  \acro{ACADO}[ACADO]{Automatic Control and Dynamic Optimization}
  \acro{ADMM}[ADMM]{Alternating Direction Method of Multipliers}
  \acro{AHC}[AHC]{Active Heave Compensation}    
  \acro{API}[API]{Application Programming Interface}
  \acro{ASV}[ASV]{Autonomous Surface Vessel}
  \acro{BLDC}[BLDC]{Brushless Direct Current}
  \acro{COG}[COG]{Center Of Gravity}
  \acro{CPU}[CPU]{Central Processing Unit}
  \acro{CTU}[CTU]{Czech Technical University}
  \acro{DOF}[DOF]{degree of freedom}
  \acroplural{DOF}[DOFs]{degrees of freedom}
  \acro{ECI}[ECI]{Earth-centered inertial}
  \acro{ECEF}[ECEF]{Earth-centered Earth-fixed}
  \acro{EKF}[EKF]{Extended Kalman Filter}
  \acro{ENU}[ENU]{East North Up}
  \acro{ESC}[ESC]{Electronic Speed Controller}
  \acro{FAA}[FAA]{Forcing Attitude Alignment}
  \acro{FFT}[FFT]{Fast Fourier Transformation}
  \acro{FOV}[FOV]{Field of View}
  \acro{GLF}[GLF]{Generalised Logistic Function}
  \acro{GNSS}[GNSS]{Global Navigation Satellite System}
  \acro{GPS}[GPS]{Global Positioning System}
  \acro{GPU}[GPU]{Graphics Processing Unit}
  \acro{HW}[HW]{Hardware}
  \acro{IBVS}[IBVS]{Image-Based Visual Servoing}
  \acro{IEKF}[IEKF]{Iterated Extended Kalman Filter}
  \acro{INS}[INS]{Inertial Navigation System}
  \acro{IMU}[IMU]{Inertial Measurement Unit}
  \acro{KF}[KF]{Kalman Filter}
  \acro{LED}[LED]{Light-Emitting Diode}
  \acro{LKF}[LKF]{Linear Kalman Filter}
  \acro{LMPC}[LMPC]{Linear Model Predictive Control}
  \acro{LPI}[LPI]{Landing Period Indicator}
  \acro{LQR}[LQR]{Linear Quadratic Regulator}
  \acro{LTI}[LTI]{Linear time-invariant}
  \acro{lidar}[LiDAR]{Light Detection and Ranging}
  \acro{MAV}[MAV]{Micro Aerial Vehicle}
  \acro{MEMS}[MEMS]{Micro Electronic Mechanical Systems}
  \acro{MOI}[MOI]{Moment Of Inertia}
  \acro{MPC}[MPC]{Model Predictive Control}
  \acro{MRS}[MRS]{Multi-robot Systems}
  \acro{NED}[NED]{North East Down}
  \acro{NLMPC}[NMPC]{Nonlinear Model Predictive Control}
  \acro{OCP}[OCP]{Optimal Control Problem}
  \acro{ODE}[ODE]{Ordinary Differential Equation}
  \acro{OpenCV}[OpenCV]{Open Source Computer Vision Library}
  \acro{OSQP}[OSQP]{Operator Splitting Quadratic Program}
  \acro{PBVS}[PBVS]{Position-Based visual servoing}
  \acro{PID}[PID]{Proportional-Integral-Derivative}
  \acro{PWM}[PWM]{Pulse Width Modulation}
  \acro{QP}[QP]{Quadratic Programming}
  \acro{RMSE}[RMSE]{Root Mean Square Error}
  \acro{ROS}[ROS]{Robot Operating System}
  \acro{RPM}[RPM]{Revolutions Per Minute}
  \acro{RTK}[RTK]{Real-time Kinematics}
  \acro{SAR}[SAR]{Search and Rescue}
  \acro{SPA}[SPA]{Signal Prediction Algorithm}
  \acro{SLAM}[SLAM]{Simultaneous Localization And Mapping}
  \acro{UAV}[UAV]{Unmanned Aerial Vehicle}
  \acro{UGV}[UGV]{Unmanned Ground Vehicle}
  \acro{UKF}[UKF]{Unscented Kalman Filter}
  \acro{USV}[USV]{Unmanned Surface Vehicle}
  \acro{UV}[UV]{UltraViolet}
  \acro{UVDAR}[UVDAR]{UltraViolet Direction And Ranging}
  \acro{VRX}[VRX]{Virtual RobotX}
  \acro{VSL}[VSL]{Visual Slide Landing}
  \acro{WAMV}[WAM-V]{Wave Adaptive Modular Vessel}
\end{acronym}

\end{document}